\documentclass[journal]{IEEEtran}
\usepackage{xcolor}

\usepackage{cite}
\usepackage{subcaption}
\ifCLASSINFOpdf
    \usepackage[pdftex]{graphicx}
    \graphicspath{{/miscfigs/}}
    \DeclareGraphicsExtensions{.pdf,.jpeg,.png}
\else
\fi
\usepackage{amsmath,amssymb,amsfonts}
\DeclareMathOperator{\Update}{Update}
\DeclareMathOperator{\MetaUpdate}{Meta-Update}
\newcommand{\siglet}[1]{#1_{\leq T}}

\newcommand{\pphi}{p_{\phi}}

\usepackage{algorithm}
\usepackage{algpseudocode}

\begin{document}
\title{Spiking Generative Adversarial Networks With a Neural Network Discriminator: Local Training, Bayesian Models, and Continual Meta-Learning}

\author{Bleema Rosenfeld,
        ~Osvaldo~Simeone, and Bipin~Rajendran% <-this % stops a space
\thanks{Osvaldo Simeone's work was supported by the European Research Council (ERC) under the European Union’s
Horizon 2020 research and innovation programme (grant agreement No. 725731). The work has also received support by Intel Labs
via the Intel Neuromorphic Research Community. Bleema Rosenfeld's work has been supported by the U.S. National Science Foundation (grant No. 1710009)}% <-this % stops a space
}

\maketitle

\begin{abstract}
Neuromorphic data carries information in spatio-temporal patterns encoded by spikes. Accordingly, a central problem in neuromorphic computing is training spiking neural networks (SNNs) to reproduce spatio-temporal spiking patterns in response to given spiking stimuli. Most existing approaches model the input-output behavior of an SNN in a deterministic fashion by assigning each input to a specific desired output spiking sequence. In contrast, in order to fully leverage the time-encoding capacity of spikes, this work proposes to train SNNs so as to match \emph{distributions} of spiking signals rather than individual spiking signals. To this end, the paper introduces a novel hybrid architecture comprising a conditional generator, implemented via an SNN, and a discriminator, implemented by a conventional artificial neural network (ANN). The role of the ANN is to provide feedback during training to the SNN within an adversarial iterative learning strategy that follows the principle of generative adversarial network (GANs). In order to better capture multi-modal spatio-temporal distribution, the proposed approach -- termed SpikeGAN -- is further extended to support Bayesian learning of the generator's weight. Finally, settings with time-varying statistics are addressed by proposing an online meta-learning variant of SpikeGAN. Experiments bring insights into the merits of the proposed approach as compared to existing solutions based on (static) belief networks and maximum likelihood (or empirical risk minimization).
\end{abstract}


% \fi
\IEEEpeerreviewmaketitle

\begin{figure}[t!]
    \centering
    \includegraphics[width=\linewidth, trim={0, 6.7in, 1.1in, 0.05in}, clip]{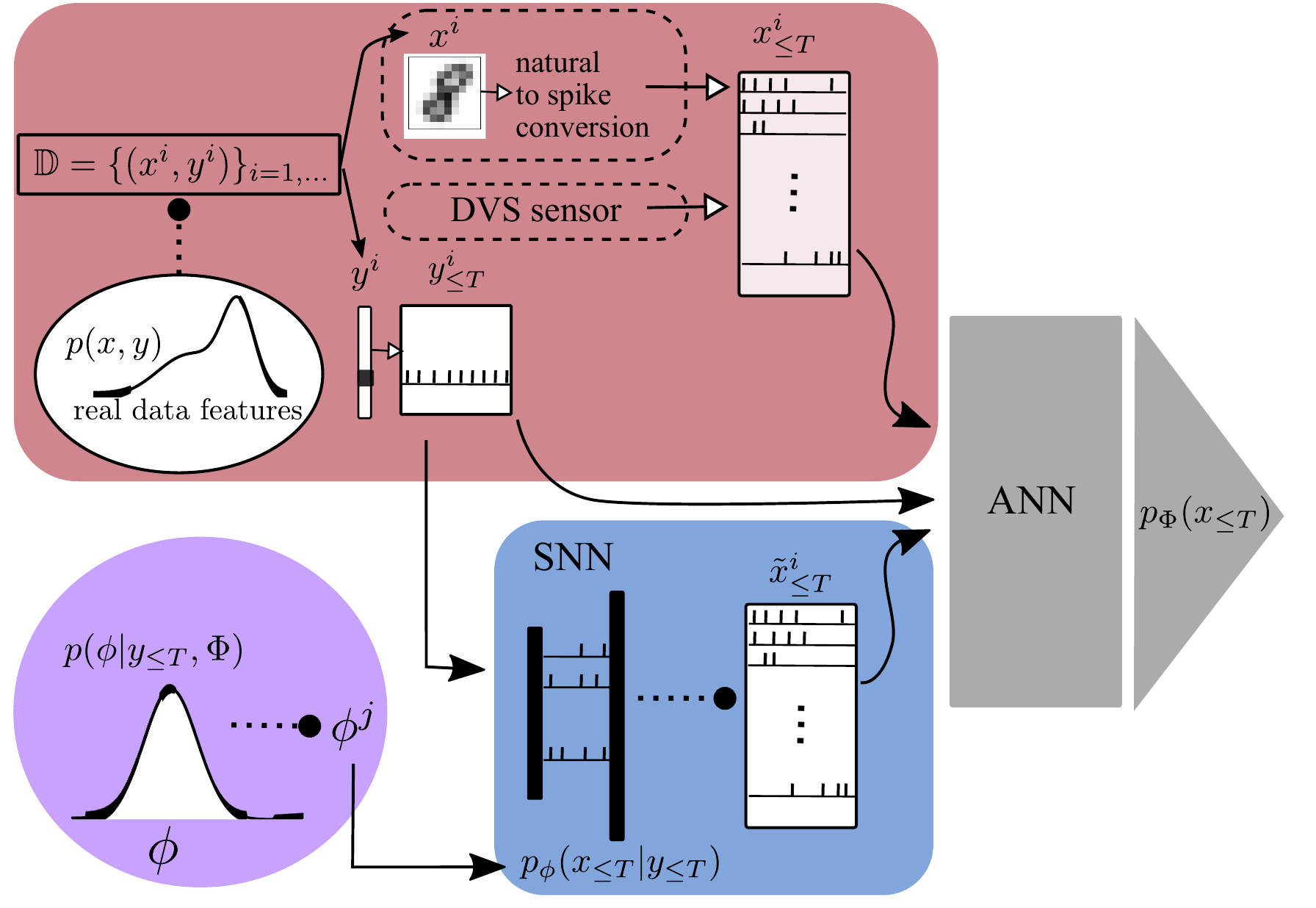}
    \caption{Block diagram of the proposed hybrid SNN-ANN SpikeGAN architecture. Sampling from the data set to obtain example $(x^i_{\leq T}, y^i_{\leq T})$ is indicated by dotted lines in the red box, along with a fixed conversion strategy from natural to spiking signals for the case of natural real data. For neuromorphic data (e.g., collected from a DVS sensor), there is no need for natural-to-spike conversion. The SNN generator, shown in the blue box, produces a sample $\tilde{x}^i_{\leq T}$. Real data and synthetic data are processed by an ANN discriminator that evaluates the likelihood $p_{\Phi}(x_{\leq T})$ that the input data is from the real data. In the case of Bayes-SpikeGAN, the model parameter $\phi^j$ of the generator are drawn from a posterior distribution, which is shown in the purple circle. Continual meta-learning is illustrated in Fig. 2.}
    \label{fig:spikegan}
\end{figure}
    % % % % % % % % % % % % % % % % % % % % % % % % % % % % % % % %
\begin{figure}[t!]
    \centering
    \includegraphics[width=\linewidth, trim={1in, 4.4in, 0.7in, 2.25in}, clip]{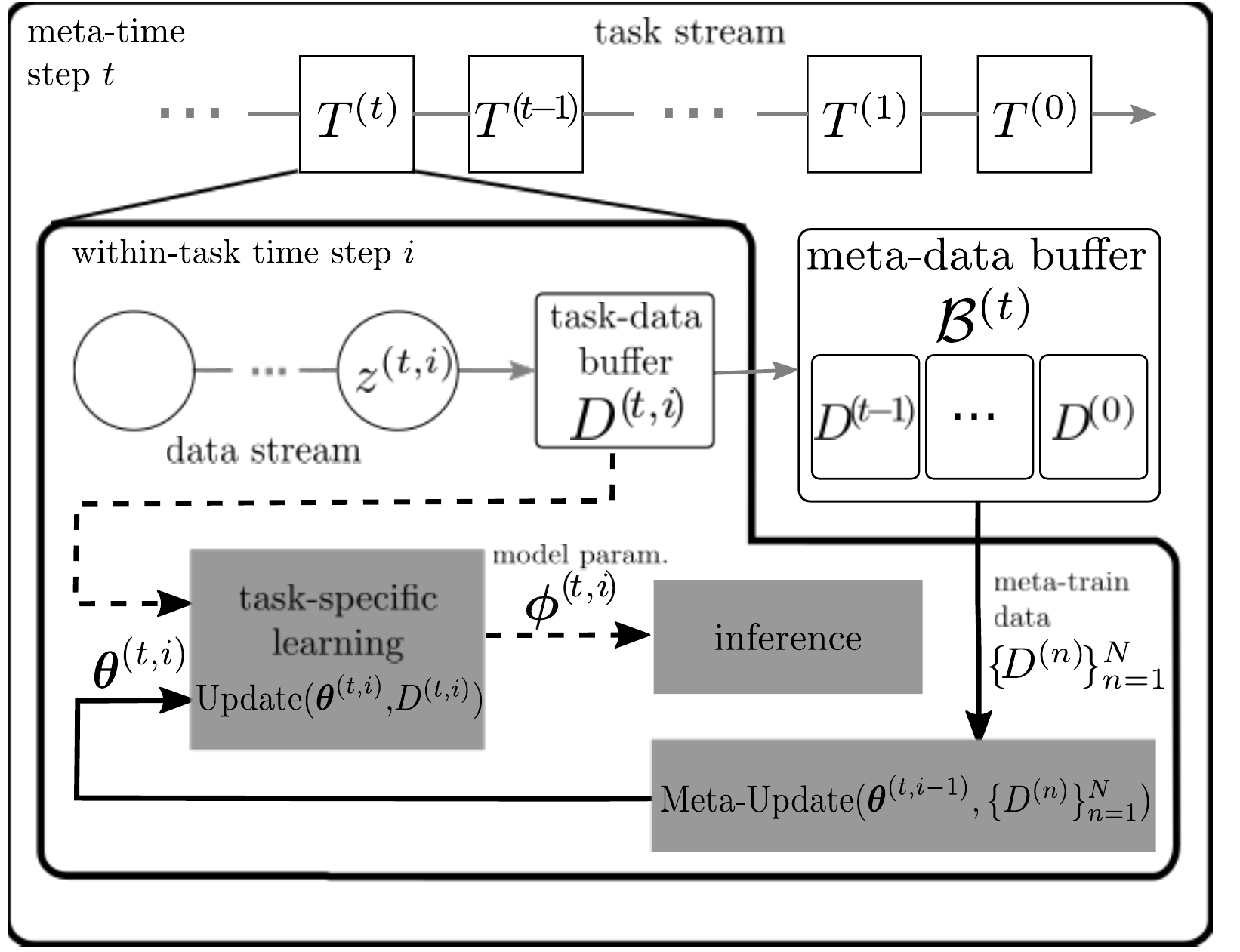}
    \caption{Online-within-online meta-learning for meta-SpikeGAN: Tasks $T^{(t)}$ are drawn from family $\mathcal{F}$ and presented sequentially to the meta-learner over timescale $t$ (denoted in the top left corner). Within-task data are also observed sequentially (bold inset box), with a new batch $z^{(t,i)}$ added to the task-data buffer $D^{(t, i)}$ at each within task time $(t, i)$. After all within-task data has been processed, the task-data buffer is added to the meta-data buffer $\mathcal{B}^{(t)}$. At each time-step $(t, i)$ the meta-learner seeks to improve online synthetic data generation by learning task-specific parameters $\boldsymbol{\phi}^{(t,i)}=\{\Phi^{(t,i)}, \phi^{(t,i)}\}$ using the updated task-data buffer and hyperparameter $\boldsymbol{\theta}^{(t, i)} = \{\Theta^{(t, i)}, \theta^{(t, i)}\}$ as the initialization (dashed arrows). Concurrently, the meta-learner makes a meta-update to the hyperparameter, yielding the next iterate $\boldsymbol{\theta}^{(t, i+1)}$. As part of the meta-update, data from $N$ different previously seen tasks are sampled from the buffer $\mathcal{B}(t)$ and task-specific parameters for $N$ parallel models are learned starting from the hyperparameter initialization $\boldsymbol{\theta}^{(t, i)}$ (solid arrows).}
    \label{fig:owometa}
\end{figure}
% % % % % % % % % % % % % % % % % % % % % % % % % % % % % % % %

\section{Introduction}

\subsection{Motivation}
The growing interest in utilizing power efficient spiking neural networks (SNNs) for machine learning tasks has spurred the development of a variety of SNN training algorithms, ranging from local approximations of backprop-through-time \cite{stewart2020chip, huh2017gradient} to local probabilistic learning rules \cite{jimenez2014stochastic, jang2019introduction}. In all of these existing solutions, inputs and outputs are prescribed spike patterns, which may be obtained from a neuromorphic data set -- e.g., from a Dynamic Vision Sensor (DVS) camera or a Dynamic Audio Sensor (DAS) recorder \cite{serrano2013128, liu2013asynchronous} -- or converted from natural signals using a fixed encoding strategies such as rate or temporal encoding \cite{pan2019neural}. This approach may lead to an overly rigid and narrow definition of the desired input-output behavior that does not fully account for the expressivity of spiking signals. Spiking signals can in fact encode the same natural signal in different ways by making use of coding in the spike times \cite{pan2019neural}. 

In this paper, we address this problem by redefining the learning problem away from the approximation of specific input-output \emph{patterns}, and towards the matching of the \emph{distribution} of the SNN outputs with a target distribution, broadening the scope of possible output spike patterns. To this end, we propose an adversarial learning rule for SNNs, and explore its extensions to a Bayesian framework as well as to meta-learning.

\subsection{Hybrid SNN-ANN Generative Adversarial Networks}

The proposed adversarial learning approach follows the general architecture of generative adversarial networks (GANs) \cite{gui2020review, saxena2021generative}.  A GAN architecture involves not only the target model, whose goal is generating samples approximately distributed according to the given desired ``real'' distribution, but also a discriminator. The role of the discriminator is to provide feedback to the generator during training concerning the discriminator's attempts to distinguish between real and synthetic samples produced by the generator. Generator and discriminator are optimized in an adversarial fashion, with the former aiming at decreasing the performance of the latter as a classifier of real against synthetic samples. 

Unlike prior work on GANs, as illustrated in Fig. 1, the key novel elements of the proposed architecture are that: (\emph{i}) the generator is a \emph{probabilistic SNN} tasked with the goal of reproducing spatio-temporal distributions in the space of spiking signals; and (\emph{ii}) the discriminator is implemented via a standard artificial neural networks (ANNs). Concerning point (\emph{i}), while in a conventional GAN model the randomness of the generator is due to the presence of stochastic, Gaussian, inputs to the generator, in this work we leverage the randomness produced by stochastic spiking neurons following the generalized linear model (GLM) \cite{pillow2008spatio, jang2019introduction}. As for the novel item (\emph{ii}), the adoption of a \emph{hybrid SNN-ANN} architecture allows us to leverage the flexibility and power of ANN-based classifiers to enhance the training of the spiking generator. It is emphasized that, once training is complete, the SNN acts as a standalone generator model, and the ANN-based discriminator can be discarded. 

Once trained, the SNN can serve as a generator model to produce synthetic spiking data with the same statistical properties of real data. This data can be used to augment neuromorphic data sets that are limited in size, or as a generative replay for SNN learning. Furthermore, since the proposed model consists of a \emph{conditional} generator, the trained SNN can also be used directly as a stochastic input-output mapping implementing supervised learning tasks without the need to hard-code spiking targets. 

As an implementation note, in neuromorphic hardware, one may envision the ANN to be implemented on the same chip as the SNN in case the deployment requires continual, on-chip, learning; or to be part of auxiliary peripheral circuitry, possibly on an external device, in case the system is to be deployed solely for inference without require continual training.

\subsection{Bayesian Spiking GANs}
The randomness entailed by the presence of probabilistic spiking neurons, in much the same way as its conventional counterpart given by Gaussian inputs, may be insufficient to produce sufficiently \emph{diverse} samples that cover real multi-modal distributions \cite{saatci2017bayesian}. A way around this problem is to allow the model parameters to be stochastic too, such that new model parameters are drawn for each sample generation. This setting that can be naturally modelled within a Bayesian framework, in which the model weights are given prior distributions that can be updated to produce posterior distributions during training as depicted by the distribution over generator weight $\phi$ in Fig. \ref{fig:spikegan}. In order to enhance the diversity of the output samples, we investigate for the first time the use of Bayesian spiking GANs, and demonstrate their potential use in reproducing biologically motivated spiking behavior \cite{weber2017capturing}.

\subsection{Continual Meta-learning for Spiking GANs}

In the continual learning setting, the statistics of the input data change over time, and a generative SNN must adapt to generate useful synthetic data. A generator model that is able to adapt based on few examples from the changed statistics is especially useful for augmenting the small data sets to be used in downstream applications such as an expert policy generator in reinforcement learning \cite{wang2020meta}. In this paper, we present a continual meta-learning framework for spiking GANs shown in Fig. \ref{fig:owometa}, in which a joint initialization is learned for the adversarial network pair made up of the SNN generator and the ANN discriminator that improves the spiking GANs efficiency in learning to generate useful synthetic data as the statistics of the population distribution vary.

\subsection{Related Work}
The only, recent, paper that has proposed a spiking GAN is \cite{kotariya2021spiking}. In it, the authors have introduced an adversarial learning rule based on temporal backpropagation with surrogate gradients by assuming a spiking generator and discriminator. The work focuses solely on encoded real-valued images, without consideration for neuromorphic data. Aside from the inclusion of Bayesian and continual meta-learning for the spiking GAN, our work explores the use of target distributions with temporal attributes, and adopts local learning rules based on probabilistic spiking models. Additionally, the probabilistic SNN model that we have adopted includes natural stochasticity that facilitates generating multi-modal synthetic data, while in \cite{kotariya2021spiking}, the randomness is artificially injected via exogenous inputs sampled from a uniform distribution and time encoded. 

Several recent works \cite{singh2020nebula, pei2019towards, yang2019dashnet, stewart2021gesture, skatchkovsky2021learning} have explored some form of hybrid SNN-ANN networks that combine SNNs and ANNs to capitalize on the low-power usage of SNNs and gain in accuracy from the broad range of processing capabilities and effective training algorithms for ANNs. Some examples of the applications are high-speed object tracking \cite{yang2019dashnet}, classification \cite{pei2019towards, singh2020nebula}, gesture recognition \cite{stewart2021gesture} and robotic control \cite{pei2019towards}. While not the main focus of our work, the SNN generator that we propose is capable of learning a temporal embedding for natural signals similar to the SNN encoder in \cite{skatchkovsky2021learning} with the key difference that in \cite{skatchkovsky2021learning} the read-out signals are a compressed encoding of real data while in our case the signals are trained to emulate the real data. Both \cite{singh2020nebula} and \cite{pei2019towards} propose chip designs to implement inference for such hybrid networks, and report increased classification accuracy for hybrid networks over pure SNN deployment with minimal increase in power usage. The Tianjic chip \cite{pei2019towards} also showcases the ability to deploy ANNs and SNNs that process simultaneously to achieve combined inference on a complex automated bicycle control task. In this case, a convolutional neural network (CNN) is used to extract features from large images while the SNN is responsible for processing auditory time series. In this work, we derive a similar benefit for the novel task of training a generative spiking model: the ANN is chosen as the discriminator to extract features of the real and synthetic data, while the SNN is responsible for modeling a temporal distribution. 

In our prior work \cite{rosenfeld2021fast}, we presented a continual meta learning framework for online SNN learning applied to classification problems. The scheme optimizes a hyperparameter initialization for the SNN that enables fast adaptation of the SNN classifier to a variety of data with similar underlying statistics. We examined its application to both natural signals under a fixed encoding method, as well as neuromorphic signals. In this work, that framework is applied to the problem of adversarial learning for SNNs to optimize a hyperparameter initialization for both the SNN generator and the ANN discriminator that enables faster optimization of the SNN generator for approximation of a range of population distributions with similar underlying statistics.

\subsection{Organization of the Paper} 
We first provide an overview of the proposed hybrid architecture and the adversarial training problem for population distributions that underlie binary temporal data in Sec. \ref{sec:spikinggan}. This is followed by a description of the SNN model that is used for the generator in Sec. \ref{sec:snnmodel}. The derivation and detailed algorithms for the proposed frequentist, and Bayesian adversarial learning rules for training a spiking generator to model a single population distribution are presented in Sec. \ref{sec:adversarial}, and Sec. \ref{sec:bayesspikegan} respectively. The continual meta-learning framework is then introduced and the algorithm for SpikeGAN training is explained in the context of adapting to multiple real distributions. The application and performance of the proposed SpikeGAN, Bayes-SpikeGAN, and meta-SpikeGAN are explored and discussed in Sec. \ref{sec:resdisc} with comparison to adversarial deep belief nets and maximum likelihood training for SNNs.  

%\IEEEpubidadjcol

\section{Hybrid SNN-ANN Spiking GAN}\label{sec:spikinggan}
In this section, we first describe the proposed hybrid SNN-ANN setting and then introduce the resulting training problem within a standard frequentist adversarial formulation.
\subsection{Setting}
In this paper, we explore adversarial training as a way to train an SNN to generate spiking signals in response to exogenous inputs whose distribution is indistinguishable from that of real data sampled conditionally from a chosen data set. As illustrated in Fig. \ref{fig:spikegan}, the proposed SpikeGAN model includes an SNN generator and an ANN discriminator. The generator outputs, termed ``synthetic data'', are processed by the discriminator along with examples from the chosen data set in an attempt to distinguish real from synthetic data. The ANN discriminator provides feedback to the SNN generator as to how realistic the synthetic samples are. This feedback is leveraged by the SNN for training which proceeds by iterating between updates to the SNN generator and the ANN discriminator.

Unlike prior work focused on the generation of static natural signals, we are concerned with generating spatio-temporal spiking data which consist of a sequence $\siglet{x} = \left(x_{i,1},...,x_{i,\tau},... x_{i,T}\right)_{i=1}^{N_x}$, of $N_x \times 1$ binary vectors $\{x_{\tau}\}_{\tau=1}^T$ over the time index $\tau=1,...,T$. The sequence is drawn from some unknown underlying population distribution $p(\siglet{x}, \siglet{y}) = p(\siglet{y})p(\siglet{x} | \siglet{y})$, jointly with another spatio-temporal spiking signal $\siglet{y} = \left(y_{i,1},...,y_{i,\tau},... y_{i,T}\right)_{i=1}^{N_y}$ with $N_y$ binary vector $\{y_{\tau}\}_{\tau=1}^T$. The signal $\siglet{y}$ may be made available to the generator $\mathcal{G}_{\phi}$ as an exogenous input to guide the generating mechanism. This auxiliary input signal can be useful to ensure that the generated data $x_{\leq T}$ cover a particular region of the data space, such as a specific class of spiking signals $\siglet{x}$.

Since the generator SNN takes as input and produces as output spiking data, in case the actual data are defined over real-valued, or non-binary discrete alphabets, an encoding, or decoding, scheme can be used to either convert between the original data format and the binary time series processed by the spiking generator, or to convert the spiking output of the generator to the format of the real data. 

The SNN generator $\mathcal{G}_{\phi}$ models the population distribution via a parameterized distribution $\pphi(\siglet{x}| \siglet{y})$ that describes the statistics of the output of the $N_x$ read out neurons given the exogenous inputs $\siglet{y}$. As detailed in Sec. \ref{sec:snnmodel}, the parameter vector $\phi$ of the SNN includes synaptic weights and biases, with the latter playing the role of firing thresholds.

The architecture of the ANN depends on whether the output of the SNN is directly fed to the ANN, which is the case when the original data are spiking signals, or if it is first converted back to a natural signal which is the case when the original data is static. The discriminator implements a classifier $\mathcal{D}_{\Phi}(\siglet{x})=p_{\Phi}(\xi=1 | \siglet{x},\siglet{y})$ giving the probability that the input $(\siglet{x}, \siglet{y})$ is drawn from the real data distribution -- an event indicated by the binary variable $\xi = 1$. 

\subsection{Training Problem}

During training, real data pairs $(\siglet{x}, \siglet{y})$ are sampled from the data set. Recall that these are spiking signals, possibly converted from natural signals. The real data pair $(\siglet{x}, \siglet{y})$ along with a synthetic data pair $(\tilde{x}_{\leq T}, \siglet{y})$, where $\tilde{x}_{\leq T}$ is the output of the generator SNN in responst to input $y_{\leq T}$, are then fed as inputs to train the ANN discriminator $\mathcal{D}_{\Phi}$. Specifically, during training, the pair of SNN and ANN models are optimized jointly, with the discriminator's parameters $\Phi$ trained to classify between the real and synthetic data, while the generator's parameters $\phi$ are updated to undermine the classification at the ANN. 

Let us denote as $\siglet{z} = (\siglet{x}, \siglet{y}) \sim p(\siglet{x}, \siglet{y})$ an input-output pair drawn from the underlying population distribution and $\siglet{z} = (\tilde{x}_{\leq T}, \siglet{y})  \sim \pphi(\tilde{x}_{\leq T}, \siglet{y})$ an input-output pair with $\siglet{y} \sim p(\siglet{y})$, drawn from the marginal population distribution, and $\tilde{x}_{\leq T} \sim \pphi(\tilde{x}_{\leq T}|\siglet{y})$ obtained from the SNN generator. In the single task frequentist setting studied in Sec. \ref{sec:adversarial}, the adversarial training objective is described by the min-max optimization problem \cite{goodfellow2014generative}
\begin{equation}\label{eq:minmax}
\min_{\phi} 
	\max_{\Phi}
			\mathbb{E}_{z_{\leq T} \sim p} \mkern-5mu\left[\!
			 \psi_1 \!\left( \!\mathcal{D}_{\Phi}(\siglet{z}\!) \!\right)\!
		\right] \!+\!
		\mathbb{E}_{z_{\leq T} \sim \pphi} \mkern-5mu\left[
			 \psi_2 \mkern-5mu \left(\! \mathcal{D}_{\Phi}(\!\siglet{z}\!) \!\right)
		\right]\!,
\end{equation} where $\psi_1(\cdot)$ is an increasing function and $\psi_2(\cdot)$ is a decreasing function. In (\ref{eq:minmax}), the first expectation is over real data sampled from the true population distribution, while the second expectation is over synthetic data generated by the SNN. Following the original GAN formulation \cite{goodfellow2014generative}, we set $\psi_1 = \log(x)$ and $\psi_2=\log(1-x)$ so that the inner maximization in (\ref{eq:minmax}) evaluates to the Jensen-Shannon divergence between the two distributions when no constraints are imposed on the discriminator. 

As we detail in the next section, in the proposed solution, stochastic gradient updates are made in a parallel fashion with the discriminator taking a gradient step to address the inner maximization in (\ref{eq:minmax}), and the generator taking a gradient step to tackle the outer minimization in (\ref{eq:minmax}). 

In Sec. \ref{sec:bayesspikegan}, the problem (\ref{eq:minmax}) is generalized to account for Bayesian SNNs in which the weight vector $\phi$ is allowed to have a posterior distribution, so that sample generation entails a preliminary step of sampling from the weight distribution \cite{saatci2017bayesian}. Furthermore, in Sec. \ref{sec:continual}, the framework described in this section is extended to continual meta-learning, in which case the population distribution varies over time.

%%%%%%%%%%%%%%%%%%%%%%%%%%%%%%%%%%%%%%%%%%%%%%%%%%%%%%%%%%%%%%%%%%%%%%%%%%%%%%%%%%%%%%
\section{Probabilistic Spiking Neuronal Network Model}\label{sec:snnmodel}
% % % % % % % % % % % % % % % % % % % % % % % % % % % % % % % % % % % % % % % %
\begin{figure}[t!]
    \centering
    \includegraphics[width=\linewidth, trim={1.4in, 6.3in, .5in, 1.9in}, clip]{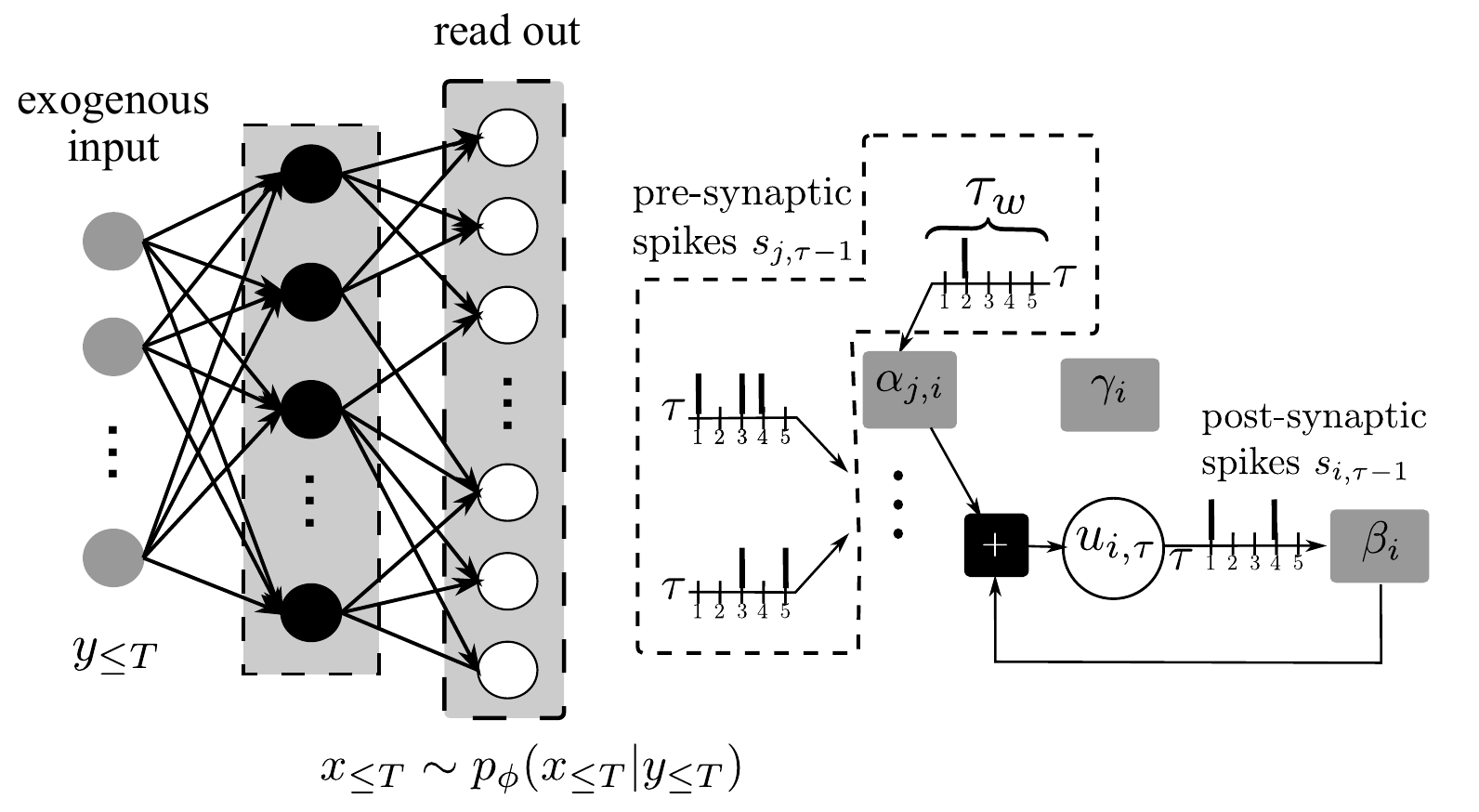}
    \caption{(\emph{Left}) An example of a generator SNN $\mathcal{G}_{\phi}$ with a fully connected topology. Black circles are the hidden neurons, in set $\mathcal{H}$, and white circles are the read-out neurons in set $\mathcal{R}$, while gray circles represent exogenous inputs. Synaptic links are shown as directed arrows, with the post-synaptic spikes of the source neuron being integrated as inputs to the destination neuron. (\emph{Right}) A pictorial representation of a GLM neuron.}
    \label{fig:snn}
\end{figure}
% % % % % % % % % % % % % % % % % % % % % % % % % % % % % % % % % % % % % % % %
In this section we describe the conditional probability distribution $\pphi(\siglet{x}| \siglet{y})$ that is followed by the samples generated by the SNN. As illustrated in Fig. \ref{fig:snn}, this distribution is realized by a general SNN architecture with probabilistic spiking neurons implementing generalized linear models (GLM) \cite{pillow2008spatio, jimenez2014stochastic,  jang2019introduction}. The generator SNN $\mathcal{G}_{\phi}$ processes data in the form of binary signals (spikes) over processing time $\tau=1,2,...$, with each neuron $i$ producing an output spike, $s_{i,\tau}=1$, or no output spike, $s_{i,\tau}=0$, at any time $\tau$. The network includes a layer of read-out neurons $\mathcal{R}$ and a set of hidden neurons $\mathcal{H}$, with respective spiking outputs denoted as $s_{i, \tau} = x_{i, \tau}, \ i \in \mathcal{R}$ and $s_{i, \tau} = h_{i, \tau},  \ i \in \mathcal{H}$.  

As depicted in Fig. \ref{fig:snn}, each neuron includes a set of pre-synaptic connections, by which signals from exogenous inputs and other neurons in the network are passed to it, as well as an auto-feedback connection through which the neuron processes its own previous outputs. The topology of the SNN is defined so as not to include any loops except for the individual neuron feedback signals. Pre-synaptic and feedback spikes are integrated via time domain filtering to update the membrane potential at every time $\tau$, giving the instantaneous membrane potential for neuron $i$
\begin{equation}\label{eq:glm_mempot}
    u_{i,\tau} = \sum_{j} \alpha_{j,i}*s_{j,\tau-1} + \beta_i*{s}_{i,\tau-1} + \gamma_i,
\end{equation} where $\alpha$ and $\beta$ are trainable pre-synaptic and feedback filters, respectively, and $\gamma_i$ is a trainable bias. Each filter $\alpha_{j,i}$ defines the pre-synaptic connection of neuron $i$ that receives inputs from neuron $j$. The expression $\alpha_{j,i}*s_{j, \tau-1}$ denotes the convolution of filter $\alpha_{j,i} $ over a window of $\tau_w$ previous signals passed through that connection. The filter is defined as a linear combination of a set of $K_a$ basis functions collected as columns of matrix $A$, such that we have ${\alpha_{j,i} = A w^{\alpha}_{j,i}}$ where $w^{\alpha}_{j,i}$ is a $K_a \times 1$ vector of trainable synaptic weights \cite{pillow2008spatio}. The post-synaptic feedback filter $\beta_{i} = B w^{\beta}_i$ is defined similarly. We collect in vector $\phi=\{w^{\alpha}, w^{\beta}, \gamma \}$ all the model parameters. 

In the GLM neuron, a post-synaptic sample $s_{i,\tau}$ is a random variable whose probability is dependent on the spikes integrated by that neuron, including hidden and read-out spiking signals. It is defined as a probabilistic function of the neuron's membrane potential $u_{i,\tau}$ at that time as
\begin{equation}\label{eq:glm_spikeprob}
     p_{\phi}(s_{i,\tau} | s_{\leq \tau}) = p_{\phi}(s_{i,\tau}=1| u_{i, \tau}) = \sigma(u_{i,\tau}),  
\end{equation} 
where we have $s_{\leq \tau} = \{h_{\leq \tau}, x_{\leq \tau}\}$,  $\sigma(x) = (1+e^{-x})^{-1}$ is the sigmoid function, and $\phi$ is the vector of trainable model parameters. 

With this expression of the output, the membrane potentials of the read-out neurons and the hidden neurons define the joint likelihood of a sequence of read-out spikes $x_{\leq \mathcal{T}} = \{[x_{i, 0},...,x_{i, \tau},...,x_{i, \mathcal{T}}]\}_{i \in \mathcal{R}}$ and hidden spikes $h_{\leq \mathcal{T}}=\{[h_{i,0},..., h_{i, \tau},..., h_{i,\mathcal{T}}]\}_{i\in\mathcal{H}}$. These sequences are sampled as a result of exogenous input sequence $y_{\leq T}$. Accordingly, the likelihood of the sequence of read-out spikes $x_{\leq \mathcal{T}}$ is conditioned on some sequence of exogenous input spikes $y_{\leq \mathcal{T}}$ and is defined as
\begin{align}\label{eq:glm_jointprob}
    \pphi(x_{\leq T}, h_{\leq T}|y_{\leq T}) & = \prod_{t=1}^T \prod_{i\in\{\mathcal{R},\mathcal{H}\}} p_{\phi_i}(s_{i,t} | u_{i,t}) \nonumber\\ & = \prod_{t=1}^T \prod_{i\in\{\mathcal{R},\mathcal{H}\}} \sigma (u_{i,t}).
\end{align} 
where $s_{i,t}$ refers to either a hidden spike signal $h_{i,t}$ or a read-out spike signal $x_{i, t}$ depending on which set the  neuron $i$ that it is sampled from belongs to.

The gradient of the log likelihood of neuron outputs  is central to the optimization and is derived as in \cite{jang2019introduction}  
\begin{align}\label{eq:grad_local}
    & \nabla_{w^{\alpha}_{j,i}} \log p_{\theta_i}(\upsilon_{i,\tau}| u_{i,\tau}) = A^T\Vec{s}_{j,\tau-1}(\upsilon_{i,\tau}-\sigma(u_{i,\tau}))\nonumber \\
    & \nabla_{w^{\beta}_i} \log p_{\theta_i}(\upsilon_{i,\tau}| u_{i,\tau}) = B^T\Vec{s}_{i,\tau-1}(\upsilon_{i,\tau}-\sigma(u_{i,\tau}))\\
    & \nabla_{\gamma_i} \log p_{\theta_i}(\upsilon_{i,\tau}| u_{i,\tau}) = (\upsilon_{i,\tau}-\sigma(u_{i,\tau})). \nonumber
\end{align}
These derivatives include a post-synaptic error term $(\upsilon_{i,\tau} - \sigma(u_{i,\tau}))$ and a pre-synaptic term $A^T\Vec{s}_{j, \tau-1}$, where $\Vec{s}_{j,\tau-1} = [s_{j, \tau-1}, s_{j, \tau-2},...,s_{j,\tau-\tau_w}]^T$ is the $\tau_w \times 1$ window of pre-synaptic spikes that were processed at time $\tau$ and $A$ is the ${\tau_w\times K_a}$ matrix of basis vectors that define the pre-synaptic filter.

%%%%%%%%%%%%%%%%%%%%%%%%%%%%%%%%%%%%%%%%%%%%%%%%%%%%%%%%%%%%%%%%%%%%%%%%%%%%%%%%%%%%

\section{Adversarial Training for SNN: SpikeGAN}\label{sec:adversarial}

As described in Sec. \ref{sec:adversarial}, the proposed SpikeGAN model for adversarial training includes an SNN generator $\mathcal{G}_{\phi}$ with parameter vector $\phi$ and an ANN discriminator $\mathcal{D}_{\Phi}$ with parameter vector $\Phi$. As illustrated in Fig. \ref{fig:spikegan}, the SNN generator processes exogenous inputs $\siglet{y}$ in a sequential manner, mapping each $N_y \times 1$ input vector $y_{\tau}$ at time $\tau$ to an $N_x \times 1$ output vector $x_{\tau}$ for $\tau = 1, ..., T$. The SNN mapping is causal and probabilistic, with an output distribution $p_{\phi}(\siglet{x}| \siglet{y})$ defined in Sec. \ref{sec:snnmodel}. The discriminator $\mathcal{D}_{\Phi}$ is implemented as an ANN with a binary classification output. In this section we propose a method, referred to as SpikeGAN, to address the training problem (\ref{eq:minmax}).
\begin{algorithm}[t]
\caption{SpikeGAN}
\label{alg:spikegan}
\begin{algorithmic}[1]
\Require Data set $\mathbb{D}=\{(x^i_{\leq T}, y^i_{\leq T})\}_{i=1,2,...}$, learning rates $\mu_{\Phi}, \mu_{\phi}$
\Repeat
\State sample a batch of real data samples $X=[x^i_{\leq T}]_{i=1}^B$ from the data set $\mathbb{D}$
\State initialize synthetic data cache $\tilde{X} = \emptyset$
\State initialize generator gradient cache $g = \emptyset$
\For{$i=1,...,B$}
	\State $x^i_{\leq T}, g^i_{\phi} \leftarrow \text{SNN procedure (see below)}$
	\State cache sample $\tilde{X} = \tilde{X} \cup \{x^i_{\leq T}\}$
	\State cache gradients $g = g \cup \{g^i_{\phi}\}$
\EndFor 
\State evaluate classification probability $\mathcal{D}_{\Phi}(X)$ and $\mathcal{D}_{\Phi}(\tilde{X})$
\State evaluate reward signal $r = [\psi_2 \left( \mathcal{D}_{\Phi} (x^i_{\leq T}) \right)]_ {i=1,...,B}$
\State Update $\Phi := \Phi + \mu_{\Phi} \frac{1}{B} \sum_{i=1}^{B} 
	\nabla_{\Phi} \psi_1 \left( \mathcal{D}_{\Phi}(x^i_{\leq T}) \right) +
	\nabla_{\Phi} \psi_2 \left( \mathcal{D}_{\Phi}(\tilde{x}^i_{\leq T}) \right)$

\State Update $\phi := \phi - \mu_{\phi} \frac{1}{B} \sum_{i=1}^B r^i g^i_{\phi}$
\Until{convergence}
\Procedure{SNN}{}
	\State initialize traces $h_{j,\tau} = 0$ for all neurons $j$ at time $\tau=1$
	\State initialize gradients $g_{\phi_j} = 0$ for all pre-synaptic connection weights to neuron $j$
	\For{$\tau=1,...,T$}
		\For{$j\in\mathcal{H}$ in order of connectivity}
			\State compute $u_{j,\tau}$ according to (\ref{eq:glm_mempot})
			\State sample $h_{j,\tau} = (h_{s,j,\tau}, h_{r,j,\tau})$ 
			\State accumulate gradients ${g_{\phi_j}\mkern-10mu \mathrel{+}=\mkern-5mu\nabla_{\phi_j}\mkern-2mu p(h_{j,\tau}\!|\mkern-1mu u_{j,\tau}\!)}$ (\ref{eq:grad_local})
		\EndFor	
	\EndFor
	\State \textbf{return} $h_{r,\leq T}, g_{\phi}$
\EndProcedure
\end{algorithmic}
\end{algorithm}
\subsection{Algorithm Overview}
Consider the population distribution $p(\siglet{x}, \siglet{y})$ with side information $\siglet{y}$ and data $\siglet{x}$ that underlies the generation of a data set $\mathbb{D}$. At each training step, a batch of $B$ examples $(x^i_{\leq T}, y^i_{\leq T})$, for $i=1,...,B$ are drawn from the data set $\mathbb{D}$ for training. Additionally, a batch of synthetic data $\tilde{x}_{\leq T}^i$ is sampled from the spiking generator as the output spikes of the $N_x$ read-out neurons given the corresponding $N_y$ exogenous inputs $y^i_{\leq T}$, for $i=1,...,B$. 

The SNN operates on the local discrete time scale defined by index $\tau=1,...,T$, which runs over the temporal dimension of each exogenous input sequence $\siglet{y}$. At each time $\tau$ it processes a batch of $B$ input vectors $y^i_{\tau}$ with $i=1,...,B$, mapping them in parallel through the full network topology to sample a batch of $B$ corresponding output vectors $\tilde{x}^i_{\tau}$, with  $i=1,...,B$ from the $N_x$ read-out neurons. As detailed in Sec. \ref{sec:snnmodel}, this involves computing batches of instantaneous membrane potentials $u_{j,\tau}$ using (\ref{eq:glm_mempot}) and sampling output spikes using (\ref{eq:glm_spikeprob}) by following the order defined by the underlying computational graph. 

 The gradients in (\ref{eq:grad_local}) for the learning criterion (\ref{eq:minmax}) are computed using a \emph{local, three-factor, rule}, and accumulated as the output spikes of each neuron are sampled over time $\tau$. After the full sequence has been processed, the  local gradients $g^i_{\phi}$, with  $i=1,...,B$ are cached for use in the outer minimization in (\ref{eq:minmax}).

The discriminator processes the batch of real data examples $\{(x^i_{\leq T}, y^i_{\leq T})\}_{i=1}^B$ and the batch of synthetic data samples $\{(\tilde{x}^i_{\leq T}, y^i_{\leq T})\}_{i=1}^B$ to approximate the expectations in (\ref{eq:minmax}). For each example, the input to the discriminator includes both the data signal $\siglet{x}$ and the feature signal $\siglet{y}$. To enable the ANN to process the time series data, the series is either compressed to a fixed smaller-dimensional embedding, or the ANN includes convolutions over the time dimension to automatically optimize suitable embeddings. The  gradient of the objective function (\ref{eq:minmax}) with respect to the discriminator parameter $\Phi$ is evaluated using standard backprop. 
\subsection{Derivation}
The GAN objective (\ref{eq:minmax}) is optimized via SGD updates with respect to the discriminator parameter vector $\Phi$ and generator parameter vector $\phi$. To update the discriminator, the gradient of the expected values in equation (\ref{eq:minmax}) are estimated by the described batches of $B$ examples drawn from the training data and from the generator as
\begin{align}
\nabla_{\Phi} \mathbb{E}_{\mathrm{z}_{\leq T} \sim p} & \left[
		\psi_1 \left( \mathcal{D}_{\Phi}(\siglet{z}) \right)
	\right]  + 
	\mathbb{E}_{\mathrm{z}_{\leq T} \sim \pphi} \left[
		\psi_2 \left( \mathcal{D}_{\Phi} (\siglet{z}) \right)
	\right] \label{eq:dis_batch_grad}\\
& \approx\! \frac{1}{B} \!\sum_{i=1}^{B} 
	\nabla_{\Phi} \psi_1\! \left( \mathcal{D}_{\Phi}(z^i_{\leq T}) \right)\! +\!
	\nabla_{\Phi} \psi_2\! \left( \mathcal{D}_{\Phi}(\tilde{z}^i_{\leq T}) \right) , \nonumber
\end{align} where $z^i_{\leq T}$ is the $i$-th example sampled from the training data and $\tilde{z}^i_{\leq T}$ is the $i$-th example sampled from the generator. The derivatives are easily computed via the standard backpropagation algorithm. 

Taking the gradient of the outer expression to update the generator model, we have 
\begin{align}\label{eq:grad_gan_wrt_gen}
\nabla_{\phi}
	& \mathbb{E}_{\mathrm{z}_{\leq T} \sim p} \left[
		\psi_1 \left( \mathcal{D}_{\Phi}(\siglet{z}) \right)
	\right]\\ 
	& + \mathbb{E}_{\tilde{\mathrm{z}}_{\leq T} \sim \pphi} \!\left[
		\psi_2 \!\left(\! \mathcal{D}_{\Phi}(\tilde{z}_{\leq T}) \right)
	\right] 
\!=\! \nabla_{\phi} \mathbb{E}_{\tilde{\mathrm{z}}_{\leq T} \sim \pphi} \!\left[
		\psi_2 \!\left( \!\mathcal{D}_{\Phi}(\tilde{z}_{\leq T}) \right)
	\right], \nonumber 
\end{align} where the derivative of the first term evaluates to zero \cite{goodfellow2014generative}. Gradient (\ref{eq:grad_gan_wrt_gen}) is estimated using the REINFORCE gradient by following \cite{jang2019introduction}. As we detail in the next section this yields a local, three-factor rule \cite{fremaux2016neuromodulated} using (\ref{eq:grad_local}).

In fact, the general form of this update for the synaptic weights is given as 
\begin{equation}\label{eq:snn_gen_update}
    w_{i,j} \leftarrow w_{i,j} + r^i \cdot g_{i,j}^i
\end{equation} where $r^i$ is the global reward signal for the $i$-th sample from the SNN generator and $g_{i, j}$ is the local neuron gradient that depends on the filtered input and output spikes. The key point of SpikeGAN is that the global reward signal ${r^i = \psi_2(\mathcal{D}_{\Phi}(x^i_{\leq T}))}$ is given by the classification certainty of the discriminator. This makes intuitive sense in that SNN connection strength is decreased if the generated outputs $x^i_{\leq T}$ are likely to be synthetic data according to the discriminator, and they are enforced in the opposite case. 

In specified experiments in Sec. \ref{sec:resdisc}, to avoid vanishing gradients early in training, we adopt the commonly used alternative generator optimization objective  $\max_{\phi} \mathbb{E}_{\mathrm{x}_{\leq T} \sim \pphi} \left[  \log \left( \mathcal{D}_{\Phi}(\tilde{x}_{\leq T}) \right) \right]$, in which the generator parameter $\phi$ is updated to maximize the log likelihood that the synthetic data is mis-classified as real data \cite{goodfellow2014generative}. The resulting gradient has the same general form (\ref{eq:grad_gan_wrt_gen}).
\begin{algorithm}[t]
\caption{Bayes-SpikeGAN}
\label{alg:bayesspikegan}
\begin{algorithmic}[1]
\Require Data set $\mathbb{D}=\{(x^i_{\leq T}, y^i_{\leq T})\}_{i=1,2,...}$, learning rates $\mu_{\Phi}, \mu_{\phi}$
\State initialize $J$ SNN generators $\mathcal{G}_{\phi}=\{\mathcal{G}_{\phi^j}\}_{j=1}^J$ each with parameter $\phi^j$
\State initialize CNN discriminator $D_{\Phi}$
\Repeat
\State sample a batch of real data samples $X=[x^i_{\leq T}]_{i=1}^B$ from the data set $\mathbb{D}$
\For{each SNN generator $G_{\phi^j}$}
    \State initialize synthetic data cache $\tilde{X} = \emptyset$
    \State initialize generator gradient cache $g = \emptyset$
    \For{$i=1,...,B$}
	    \State $x^i_{\leq T}, g^i_{\phi^j} \leftarrow \text{SNN procedure (see Alg. \ref{alg:spikegan})}$
	    \State cache sample $\tilde{X} = \tilde{X} \cup \{x^i_{\leq T}\}$
	    \State cache gradients $g = g \cup \{g^i_{\phi^j}\}$
    \EndFor 
    \State evaluate reward signal $r\! = \![-\!\log\!\left( \!\mathcal{D}_{\Phi} (x^i_{\leq T}) \right)]_ {i=1}^B$
    \State cache gradients w.r.t. all weights in $\phi^j$ $$\nabla_{\phi^j} \mathbb{E}_{\tilde{z}_{\leq T}\sim p_{\phi^j}}[-\log(\mathcal{D}_{\Phi}(\tilde{z}_{\leq T}))]\!= \!\frac{1}{B} \sum_{i=1}^B r^i g^i_{\phi}$$  
    \State evaluate and cache classification probability $\mathcal{D}_{\Phi}(X)$ and $\mathcal{D}_{\Phi}(\tilde{X})$
\EndFor
\For{each SNN generator parameter $\phi^j$}
    \State Update $\phi^j$ following Eq. \ref{eq:svgd_update} using cached gradients w.r.t. all generator parameters $\{\phi^j\}_{j=1}^J$ 
\EndFor
\State Update discriminator parameter: $$\Phi\! :=\! \Phi + \mu_{\Phi} \frac{1}{JB} \sum_{j=1}^{J}\mkern-6mu 
\nabla_{\Phi}\! \left[ \!\psi_1 \!\left(\! \mathcal{D}_{\Phi}(X^{j}_{\leq T})\! \right) \!+\!
 \psi_2\! \left(\! \mathcal{D}_{\Phi}(\tilde{X}^{j}_{\leq T})\! \right)\right]$$
\Until{convergence}
\end{algorithmic}
\end{algorithm}
\subsection{Comparison with Other Methods}
Maximum likelihood (ML) learning for SNNs optimizes the likelihood that the output signals of the read-out neurons match a target binary sequence. A major benefit of adversarial training for SNNs as compared to ML learning is its potential to better capture the different modes of the population distribution. In contrast, ML tends to produce inclusive approximations that overestimate the variance of the population distribution. A practical advantage of ML learning , as detailed in \cite{jang2019introduction} is that it enables online, incremental, learning. The proposed GAN training algorithm applies an episodic rule in which the global learning signal can only be evaluated after a full sequence of outputs has been sampled from the SNN due to the choice of an ANN as the discriminator. An online learning variant could be also devised by defining the discriminator as a recurrent network \cite{yoon2019time}, but we leave this topic to future research. 

\section{Bayes-SpikeGAN}\label{sec:bayesspikegan}
In the previous section, we have explored frequentist adversarial training for SNNs. By forcing the choice of a single model parameter vector $\phi$ for the generator $\mathcal{G}_{\phi}$, the approach may fail at reproducing the diversity of multi-modal population distribution \cite{saatci2017bayesian}. As an example, it has been shown that tailored synaptic filters and spiking thresholds are necessary to induce specific temporal patterns at the output of a spiking neuron \cite{weber2017capturing}, which are incompatible with the choice of a single parameter vector $\phi$. In this section we explore the application of Bayesian adversarial learning to address this problem.
\subsection{Generalized Posterior}
The Bayes-SpikeGAN assumes a prior distribution, $p(\phi)$, over the generator parameter vector $\phi$, and, rather than optimizing over a single parameter vector, it obtains a generalized posterior distribution on $\phi$ given observed real data. For a fixed ANN discriminator, the posterior distribution can be defined as \cite{saatci2017bayesian}
\begin{equation}\label{eq:post_dist_phi}
    p(\phi|y_{\leq T}, \Phi) \propto p(\phi) \mathbb{E}_{\tilde{x}_{\leq T}\sim p_{\phi}} \left[\mathcal{D}_{\Phi}(\tilde{x}_{\leq T})\right],
\end{equation} where the expectation is over synthetic data $\tilde{x}_{\leq T}$ sampled from the distribution $p_{\phi}(x_{\leq T})$ defined by the generator $\mathcal{G}_{\phi}$. In (\ref{eq:post_dist_phi}), the average  confidence of the discriminator $\mathcal{D}_{\Phi}(\tilde{x}_{\leq T})$ plays the role of likelihood of the current generator parameter $\phi$ given the observed real data used to optimize discriminator parameter vector $\Phi$. 

\subsection{Training Objective}
Computing the generalized posterior (\ref{eq:post_dist_phi}) is generally intractable, and hence we approximate it with a variational distribution $q(\phi|y_{\leq T}, \Phi)$. The variational posterior $q(\phi|y_{\leq T}, \Phi)$ is optimized by addressing the problem of minimizing the free energy metric 
\begin{equation}\label{eq:KLdiv}
    \min_{q(\phi)} -\log\left( \mathbb{E}_{\tilde{x}_{\leq T}\sim q(\phi)}[\mathcal{D}_{\Phi}(\tilde{x}_{\leq T})]\right) - \textrm{KL}(q(\phi)||p(\phi)),
\end{equation}where $\textrm{KL}(q(\phi)||p(\phi))=\mathbb{E}_{\phi \sim q(\phi)}[\log(q(\phi)/p(\phi))]$ is the Kullback-Liebler (KL) divergence. If no constraints are imposed on the distribution $q(\phi)$, the optimal solution of problem (\ref{eq:KLdiv}) is exactly (\ref{eq:post_dist_phi}). We further apply Jensen's inequality to obtain the more tractable objective \begin{equation}\label{eq:KLdivJensens}
    \min_{q(\phi)} -\mathbb{E}_{\tilde{x}_{\leq T}\sim q(\phi)}\left[\log \mathcal{D}_{\Phi}(\tilde{x}_{\leq T})\right] - \textrm{KL}(q(\phi)||p(\phi)).
\end{equation} 

In order to address this problem, we parametrize the variational posterior with a set of $J$ parameter vectors $\phi = \{\phi^j\}_{j=1}^J$, also known as particles. This effectively defines $J$ SNN generators $\{\mathcal{G}_{\phi^j}\}_{j=1}^J$. Samples from the generator are then obtained by randomly and uniformly selecting one particle from the set of $J$ particles, and then using the selected sample $\phi^j$ to run the SNN generator  $\mathcal{G}_{\phi^j}$.

As explained next, in order to optimize the set of particles with the goal of minimizing the free energy metric in (\ref{eq:KLdiv}), we leverage Stein variational gradient descent (SVGD) \cite{liu2016stein}.

\subsection{Bayes-SpikeGAN}
Following SVGD, for a fixed discriminator parameter vector $\Phi$, the particles are updated simultaneously at each iteration as 
\begin{align}\label{eq:svgd_update}
    &\phi^j_{i+1} = \phi^j_i \nonumber\\
    &- \eta \sum_{j'=1}^J \! \Big\{ \!\kappa(\phi^j_i, \phi^{j'}_i\!)\left(\!-\nabla_{\phi^{j'}}\mkern-3mu\mathbb{E}_{\tilde{x}_{\leq T} \sim p_{\phi^{j'}}}\mkern-7mu\left[\log\left( p(\phi)\mathcal{D}_{\Phi}(\tilde{x}_{\leq T}\!)\right)\!\right]\right) \nonumber \\
    &\hspace{40pt}- \nabla_{\phi^{j'}}\kappa(\phi^j_i, \phi^{j'}_i) \Big\}
\end{align} where $\kappa(\phi^j, \phi^{j'}) = \exp(-||\phi^j- \phi^{j'}||^{2})$ is the Gaussian kernel function. The gradient with respect to the generator parameter $\phi^{j'}_i$ can be computed as in  (\ref{eq:grad_gan_wrt_gen}),  which is estimated  via the REINFORCE gradient as $r^i\cdot g^i_{i, j}$ as in  (\ref{eq:snn_gen_update}). 

In each iteration, the discriminator parameter, $\Phi$, is updated via SGD to optimize the standard GAN loss function given in (\ref{eq:minmax}) by taking an average of the losses calculated for data sampled from each of the $J$ generators.

\section{Continual Meta-Learning for Spiking GANs: Meta-SpikeGAN}\label{sec:continual}
We have so far defined two adversarial training methods for SNNs, namely SpikeGAN (based on frequentist learning) and Bayes-SpikeGAN to train an SNN to generate data that follows a single population distribution. We now focus on a general continual meta-training framework that can be combined with SpikeGAN  adversarial training in order to enable the SpikeGAN to efficiently, and sequentially, learn how to generate data from a range of similar population distributions. 

Meta-learning assumes the presence of a family $\mathcal{F}$ of tasks that share common statistical properties. Specifically, it assumes that a common hyperparameter $\theta$ can be identified that yields efficient learning when applied separately for each task in $\mathcal{F}$. Following the current dominant approach \cite{finn2019online,nichol2018first}, we will take hyperparameter $\theta$ to represent the initialization to be used for the within-task training iterative procedure. This hyperparameter initialization improves the learning efficiency of the within-task training in terms of the total updates necessary to obtain a useful within-task model. 
\subsection{Problem Setting}
In the continual meta-learning formulation adapted from \cite{finn2019online}, the meta-learner improves the hyperparameter initialization over a series of tasks drawn from family $\mathcal{F}$ while simultaneously learning a task specific parameter for each task. As each new task is observed it aims to improve the across task generalization capability of the hyperparameter while maintaining the ability to quickly recover the within task parameter learned for previous tasks. To this end, the meta-learner runs an underlying meta-learning process to update the hyperparameter $\theta$ by using data observed from previous tasks in the series. The hyperparameter $\theta$ is then used as a within-task model initialization that enables efficient within-task training for the new task. 

To support these two processes, two data buffers are maintained. The \textit{task-data buffer} collects streaming within-task data used for within-task learning, While the \textit{meta-data buffer} holds data from a number of previous tasks to be used by the meta-training process. As illustrated in Fig. \ref{fig:owometa}, a stream of data sets $\mathbb{D}^{(t)}$, each corresponding to a task $T^{(t)}\in \mathcal{F}$, is presented to the meta-learner sequentially at $t=1,2,...$. Within each \textit{meta-time step} $t$, samples from data set $\mathbb{D}^{(t)}$ are also presented sequentially, so that at each \textit{within-task time step} $i$, a batch  $\mathit{z}^{(t,i)} = \{(x^j,y^j)\}_{j=1}^{B} \subseteq \mathbb{D}^{(t)}$ of $B$ training examples for task $T^{(t)}$ is observed, and added to the task-data buffer as  $D^{(t,i)} = D^{(t, i-1)} \cup \mathit{z}^{(t,i)}$ with $D^{(t,0)} = \emptyset$. Once all within-task data for task $T^{(t)}$ has been processed, the final task-data buffer $D^{(t, i)}$ is added to a meta-data buffer $\mathcal{B}^{(t)}$. 
\subsection{Algorithm Overview} 
The within-task training and meta-training processes take place concurrently at each time $(t, i)$. As a new batch of within-task data is observed for the current task $T^{(t)}$, the meta-learner uses it, along with the entire current task-data buffer $D^{(t,i)}$, to learn a better task-specific model parameter $\phi^{(t, i)}$ and thus improve the quality of generated synthetic data for that task. The task-specific parameter is initialized with the current hyperparameter $\theta^{(t, i)}$ and is updated via an iterative within task training process ${\phi^{(t, i)} = \Update(\theta^{(t, i)}, D^{(t,i)})}$. Concurrently, the meta-learner improves the hyperparameter initialization for the next round of within-task training by making a single gradient update to $\theta$. This update can be written as ${\theta^{(t,i+1)} \leftarrow \theta^{(t, i)} + \mu \nabla_{\theta}F(\theta^{(t, i)}, \mathcal{B}^{(t)})}$ for some meta-learning rate $\mu \geq 0$, where $F$ is the meta-learning objective function. The meta-learning objective evaluates the performance of the initialization $\theta^{(t, i)}$ on data from previous tasks stored in the meta-data buffer $\mathcal{B}^{(t)}$. 

Specifically, in order to evaluate the meta-learning objective function $F$, task-specific parameters for a number of previous tasks need to be learned. To this end, $N$ tasks $T^{(n)}, \ n=1,...,N,$ are drawn from the meta-data buffer $\mathcal{B}^{(t)}$ and a small data-set, $D^{(n)}$, is drawn as a subset of the stored data for each task. The within-task iterative training process is applied to learn task-specific parameters $\phi^{(n)} = \Update(\theta^{(t, i)}, D^{(n)})$ using $N$ parallel models, each initialized with the hyperparameter $\theta^{(t, i)}$. 

\begin{figure}[t!]
    \centering
    \includegraphics[width=\linewidth, trim={0.3in, 0.0in, 0.3in, 0.5in}, clip]{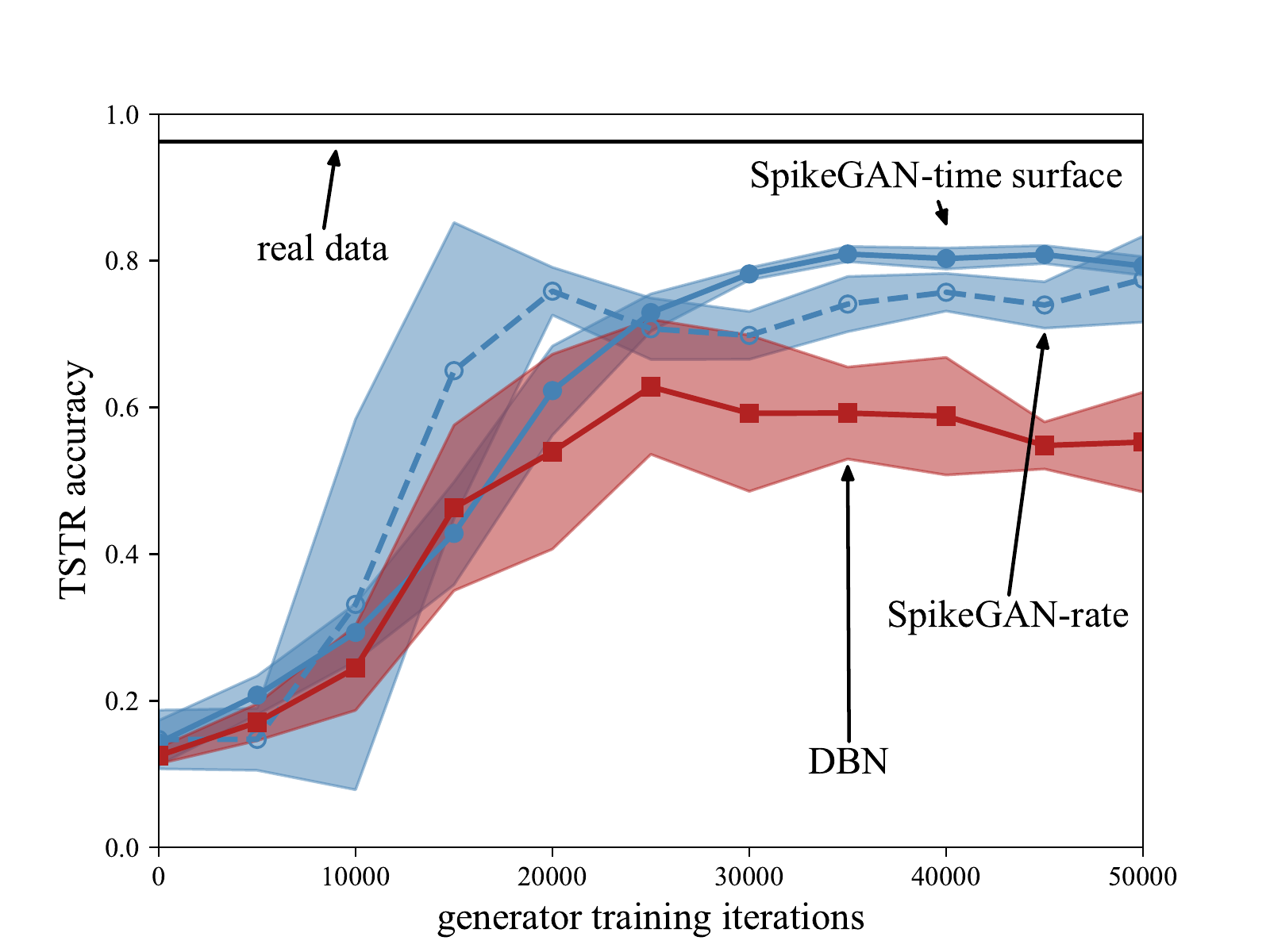}
    \caption{TSTR classification accuracy for synthetic data sampled from the SNN generator during training. The black line is the ideal test accuracy for a classifier trained with real data. The blue lines are results from SpikeGAN with outputs converted to images using rate (blue dashed) or time surface (blue solid) decoding, while the red line represents the performance of the DBN \cite{huang2019deep}.}
    \label{fig:batch100varydecodetrainfake}
\end{figure}

\subsection{Meta-SpikeGAN}
We are now ready to adapt the continual meta-learning framework to the SpikeGAN architecture described in the previous section -- a system we will refer to as meta-SpikeGAN. We start by introducing two meta-models, a discriminator ANN $\mathcal{D}_{\Theta}$, and a spiking generator $\mathcal{G}_{\theta}$, are defined, whose weights $\Theta$ and $\theta$ respectively, define the hyperparameters ${\boldsymbol{\theta}^{(t, i)} = \{\Theta^{(t,i)}, \theta^{(t,i)} \}}$ that will be updated during the meta learning process. The two hyperparameter vectors must be learned synchronously in order to maintain the balance between the discriminator and the generator in the min-max process of adversarial learning described by (\ref{eq:minmax}). In particular, it is important that the discriminator learn to differentiate real and synthetic data quickly, based on few examples from the new task, in order to provide a meaningful learning signal to the spiking generator. While the derivation here follows a frequentist framework, extensions to Bayesian solutions could be obtained by following the approach detailed in the previous section.

In meta-SpikeGAN, the within-task iterative update function $\Update(\theta, D)$ refers to the adversarial training process described in Sec. \ref{sec:adversarial} in which both models are updated to learn within task parameters $\Phi$ and $\phi$. At each within-task time-step $(t, i)$, $N+1$ adversarial model pairs are instantiated with initial weight given by $\boldsymbol{\theta}^{(t, i)}$. One pair is trained using data from the current task $T^{(t)}$ to generate within task synthetic data, while the remaining $N$ adversarial network pairs are used to enable the meta-update.

% \begin{figure}[t!]
%     \begin{minipage}{0.45 \linewidth}
%          \includegraphics[width=\linewidth]{results/pca_iter50000_dbnbinary.pdf}
%     \end{minipage}
%     \hfill
%     \begin{minipage}{0.45\linewidth}
%          \includegraphics[width=\linewidth]{results/pca_iter50000_snnnorm.pdf}
%     \end{minipage}
%     \caption{PCA visualizing the projection of the real data set into the space defined by its top two eigenvectors in red and the projection of an equivalent synthetic data set into that space in blue. Left: synthetic data sampled from the DBN generator. Right: synthetic data sampled from the spiking generator with time surface decoding (ours).}
%     \label{fig:pca_dbnsnn}
% \end{figure}

% \begin{figure}[t!]
%     \centering
%     \begin{minipage}{\linewidth}
%         \centering
%         \includegraphics[width=\linewidth, trim={0.2in, 0.5in, 0.1in, 0.7in}, clip]{results/spike_count_iter50000_timesurfacedecode.png}
%     \end{minipage}
%     \vfill
%     \begin{minipage}{\linewidth}
%         \centering
%         \includegraphics[width=\linewidth, trim={0.2in, 0.0in, 0.1in, 0.7in}, clip]{results/spike_count_iter50000_ratedecode.png}
%     \end{minipage}
%     \caption{Average number of spikes per output neuron over 100 conditional spiking GAN synthetic data samples under time surface decoding (top) and rate decoding (bottom)}
%     \label{fig:spikecount}
% \end{figure}
\begin{figure*}[t!]
    \begin{subfigure}{0.65\linewidth}
    \begin{minipage}{\linewidth}
    \centering
    \includegraphics[width=\linewidth, trim={0.0in, 0.1in, 0.0in, 0.in}, clip]{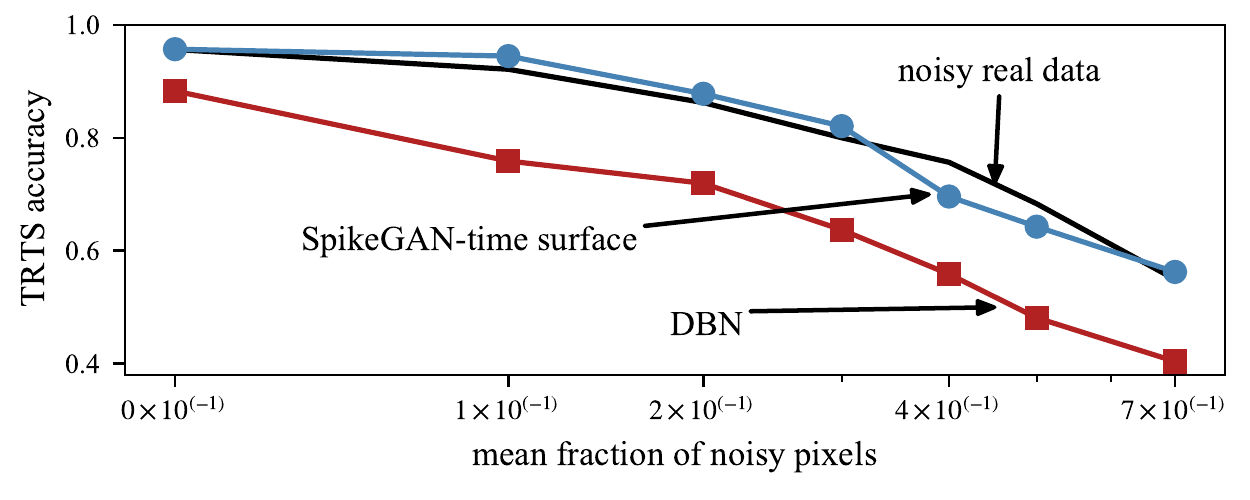}
    \caption{}
    \label{fig:noisedbnvsnn}
    \end{minipage}
    \vfill
    \begin{minipage}{\linewidth}
        \hspace{0.1in}\includegraphics[width=\linewidth, trim={0.0in, 0.0in, 0.2in, 0.05in}, clip]{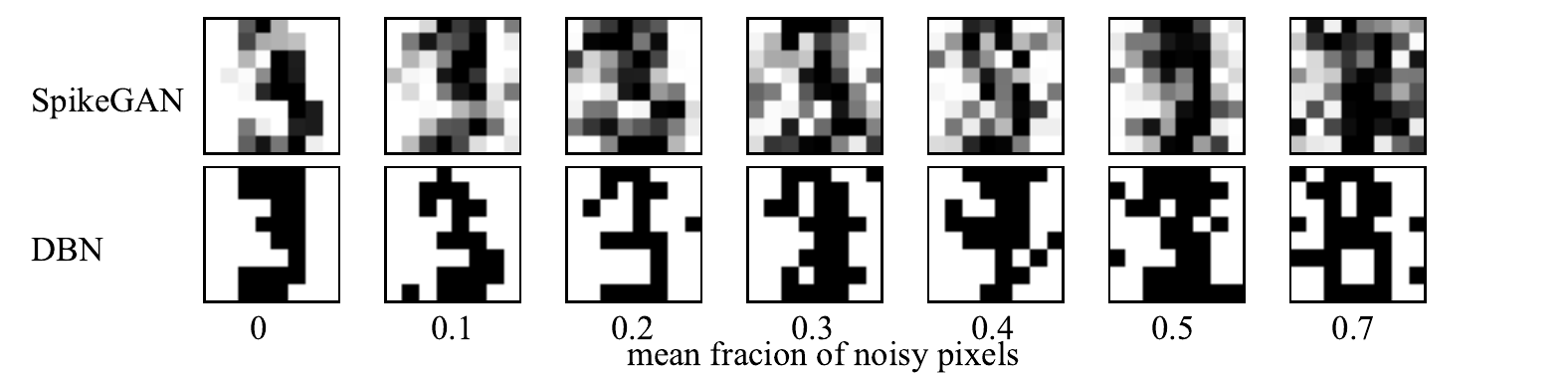}
        \caption{}
        \label{fig:noisydata_dbnsnn}
    \end{minipage}
    \end{subfigure}
    \hfill
    \begin{subfigure}{0.3\linewidth}
        \begin{minipage}{\linewidth}
        \centering
         \includegraphics[width=0.7\linewidth, trim={0.2in, 0.2in, 0.75in, 0.75in}, clip]{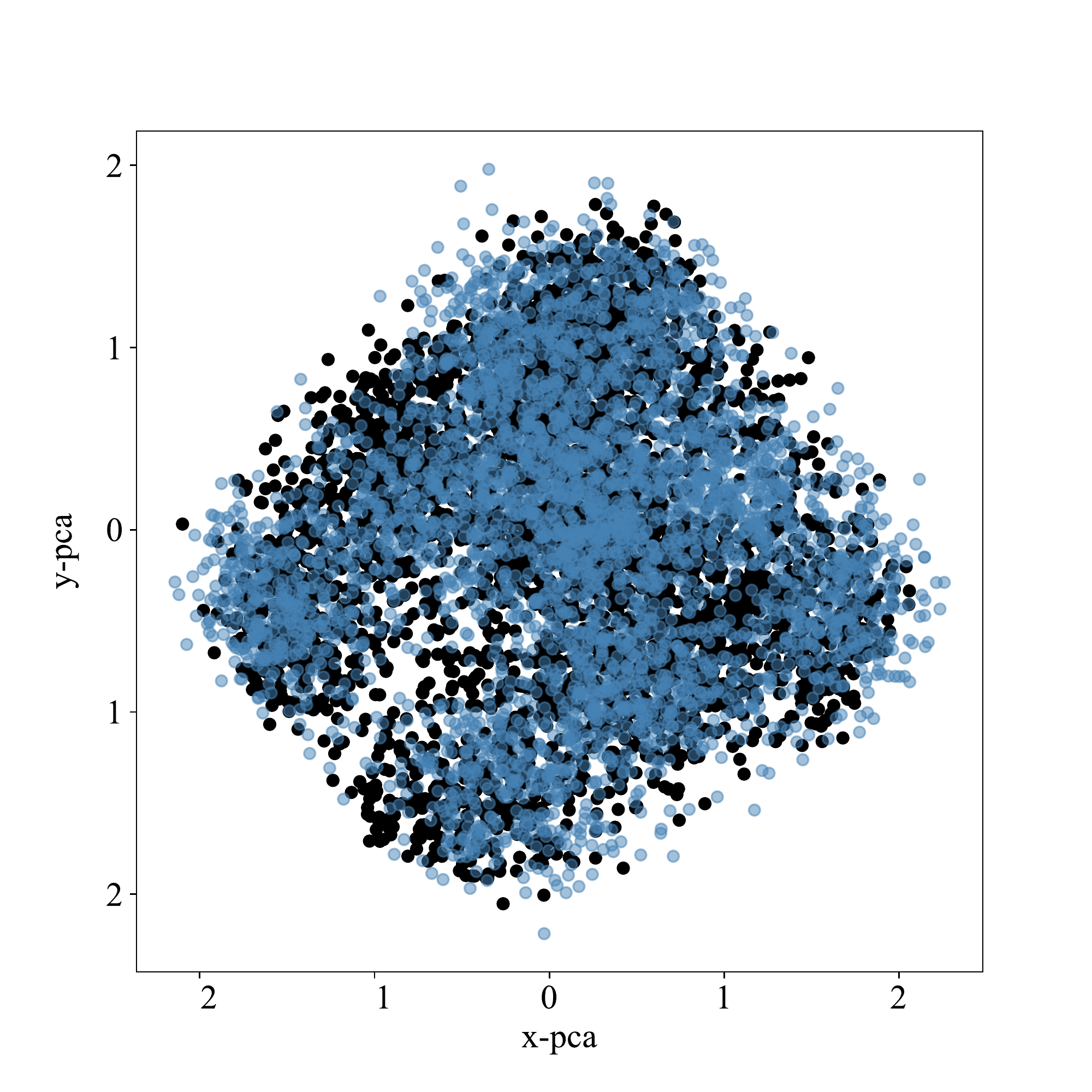}
        \caption{}
       \label{fig:pca_snntimedecode}
    \end{minipage}
    \vfill
    \begin{minipage}{\linewidth}
    \centering
        \includegraphics[width=0.7\linewidth, trim={0.3in, 0.2in, 0.75in, 0.75in}, clip]{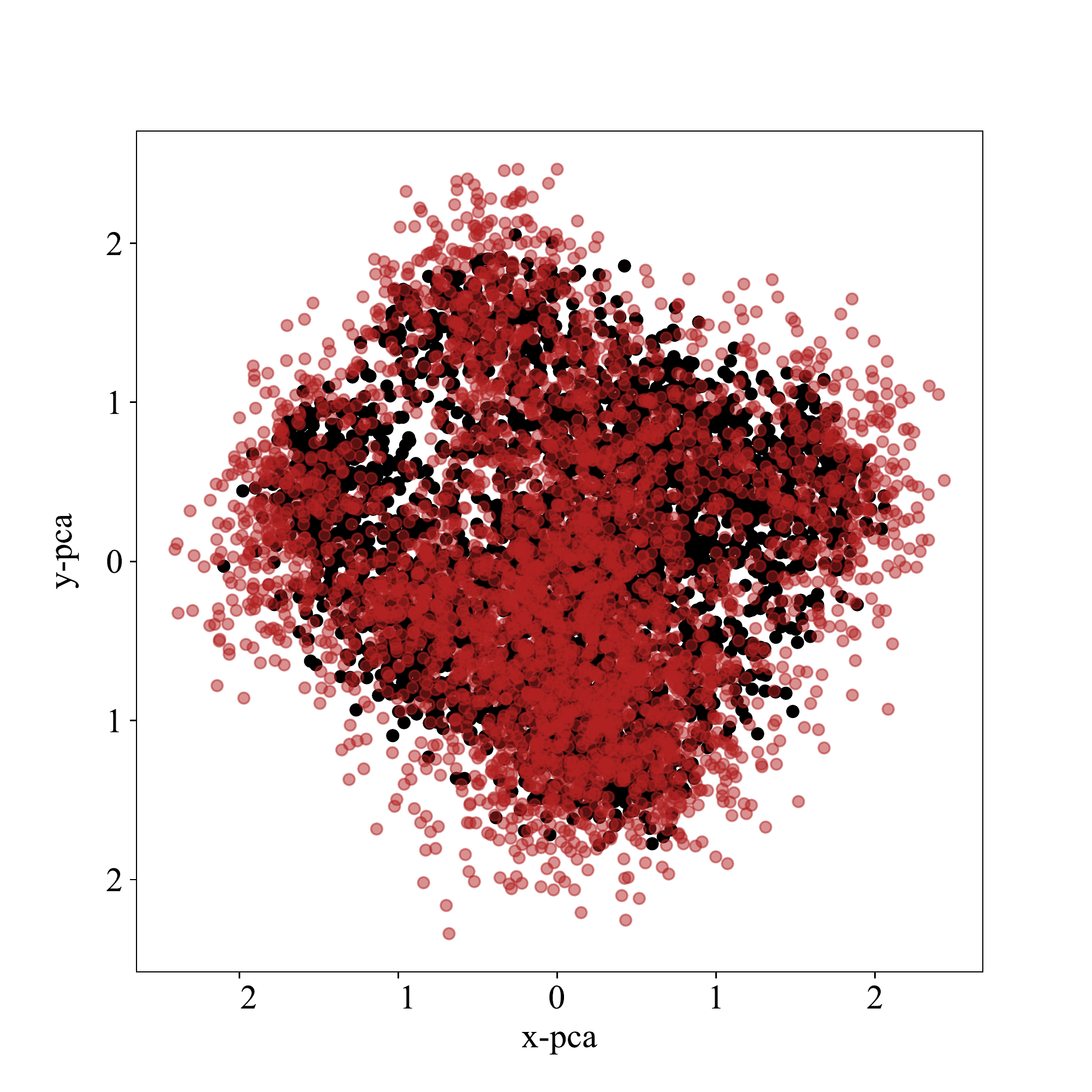}
        \caption{}
       \label{fig:pca_dbn}
    \end{minipage}
    \end{subfigure}
    \caption{(a) Handwritten digit classification accuracy for test data sampled from a generator trained using a noisy real data set. The fraction of pixels per image corrupted by additive uniform noise is increased on a $\log$ scale. Test data is sampled from either SpikeGAN (blue), DBN GAN (red) or the noisy real data set (black) as a baseline. (b) Synthetic data sampled from SpikeGAN with time surface decoding (top) and DBN GAN (bottom) trained over real data disrupted by additive uniform noise. The fraction of noisy pixels per image in the real data increases from left to right as labeled. (c,d) PCA projections of SpikeGAN synthetic data (top) and DBN GAN synthetic data (bottom) onto the real data principal components. The real data projection is shown by the black dots in both figures.}
\end{figure*}
To elaborate, at every meta-time step $t$, a task $T^{(t)}\in\mathcal{F}$ is drawn, and the task-data buffer is initialized as $D^{(t,i)} = \emptyset$. Within-task data  is added to the current task-data buffer at every within-task time step $i$ in batches of $B$ training examples $z^{(t,i)} = \{(x^j_{\leq \mathcal{T}}, y^j_{\leq \mathcal{T}})\}_{j=1}^B$. The discriminator and the spiking generator are initialized with hyperparameter $\Theta^{(t,i)}$ and $\theta^{(t,i)}$ respectively and trained via the update function $\Update(\boldsymbol{\theta}^{(t,i)}, D^{(t,i)})$ over the data in the task-data buffer.

The update function $\Update(\boldsymbol{\theta}, D)$ addresses the problem of within-task adversarial learning of model parameters $\phi$ and $\Phi$ starting from initializations $\theta$ and $\Theta$. The meta-objective, following the classic MAML formulation, is to optimize the min-max adversarial training objective over the hyperparameter initialization $\boldsymbol{\theta}$, given the learned within-task parameters across multiple previously seen tasks. Specifically, the objective is defined as an average over $N$ tasks with data sets $\mathbb{D}^{(n)}$ stored in the meta-data buffer as
\begin{align}\label{eq:meta_obj}
  \min_{\theta}\max_{\Theta}\sum_{n=1}^{N} &
  \sum_{i=1}^B \log\left(\mathcal{D}_{\Phi^{(n)}}(x^{(n),i}_{\leq T})\right) +
  \\ & +\sum_{i=1}^B \log\left(\!1\!-\!\mathcal{D}_{\Phi^{(n)}}(\tilde{x}^{(n),i}_{\leq T})\!\right)p_{\phi^{(n)}}(\tilde{x}^{(n),i}_{\leq T}|y^{(n), i}_{\leq T}) \nonumber 
\end{align} where the real data is sampled from the  data set $\mathbb{D}^{(n)}$ and the synthetic data is sampled from the generator $G_{\phi^{(n)}}$ trained via within-task adversarial training for that task.

To implement the meta-update function $\MetaUpdate\left(\boldsymbol{\theta}^{(t,i)}, \{D^{(n)}\}_{n=1}^N\right)$, the mentioned $N$ adversarial network pairs are trained in parallel using $N$ data-sets sampled from the meta-data buffer $\{D^{(n)}\}_{n=1}^N \in \mathcal{B}^{(t)}$. Each of the data-sets includes $M$ training examples from the real data that are a subset of the data set of a previously seen task such that ${D^{(n)} = \{(x^j_{\leq \mathcal{T}}, y^j_{\leq \mathcal{T}})\}_{j=1}^M}$. Once the within task parameters for both networks ($\mathcal{D}_{\Phi^{(n)}}$ and $\mathcal{G}_{\phi^{(n)}}$) for each of the $N$ tasks are learned, the hyperparameters $\Theta^{(t,i)}$ and $\theta^{(t,i)}$ are each updated individually as discussed next.

The update of the hyperparameters implemented by the meta-update function requires the computation of the second order gradients of the objective function used in the within-task learning update. In this work, we make use of the first-order REPTILE approximation for the gradient which has been shown to have properties similar to the true gradient in a number of benchmark tasks \cite{nichol2018first}. Accordingly, the hyperparameters are individually updated as
\begin{align}
    \Theta^{(t, i+1)} = \Theta^{(t, i)} - \Phi^{(n)} \label{eq:d_reptile_update}\\
    \theta^{(t, i+1)} = \theta^{(t, i)} - \phi^{(n)}\label{eq:g_reptile_update}. 
\end{align}

% \subsection{Mathematical Formulation}

% The meta-gradient $\nabla_{\theta}F (\theta, \mathcal{B}^{(t)})$ requires the computation of the second order gradient of the training losses used in the task-specific learning function $\Update(\theta, D^{(n)}_{task})$ \cite{finn2017model}. In this work, we make use of the first-order REPTILE approximation for the gradient 
% \begin{equation}\label{eq:reptile_udpate}
%     \nabla_{\theta}\log p\left( D^{(n)}_{meta} \middle| \phi^{(n)}\right) \approx
%     \theta^{(t, i)} - \phi^{(n)},
% \end{equation}
% which has been shown to have properties similar to the true gradient in a number of benchmark tasks \cite{nichol2018first}. This yields the meta-update function 
% \begin{align}
%     \MetaUpdate&(\theta^{(t,i)}, \{D^{(n)}\}_{n=1}^N) =\\ &\theta^{(t, i)} \!+ \mu \!\sum_{n=1}^N \!\left(\theta^{(t, i)} \!- \Update(\theta^{(t, i)}, D^{(n)}_{task})\right),\nonumber
% \end{align} where $\mu \geq 0$ is the meta-learning rate.
% \begin{figure*}[t!]
% \begin{subfigure}{0.55\linewidth}

\section{Experimental Methods}\label{sec:methods}
In this section, we describe the experimental set-up we have adopted to evaluate the performance of SpikeGAN, Bayes-SpikeGAN, and meta-SpikeGAN.

\subsection{Data Sets, Encoding, and Decoding} 

We consider three different data sets: 1) handwritten digits \cite{Dua:2019}; 2) simulated spike-domain handwritten digits; and 3) synthetic temporal data \cite{weber2017capturing}. These data sets have been selected to present a range of spatial and temporal correlations, posing different challenges to the training of a generative model. The handwritten digits data set represents a population distribution with exclusively spatial correlations, as there is no temporal aspect to the real data. The simulated spike domain handwritten digits data set incorporates some temporal correlations by using a spike code to convert the handwritten digit data into the spike domain. Lastly, the synthetic temporal data has a dimension of $N_x = 1$ and thus includes only strong temporal features and no spatial correlations. The data sets are detailed in the next section.

For the first data set, the SNN outputs at the read-out layer are compressed to match the domain of the real data, using rate decoding or time surface decoding \cite{stewart2021gesture}. Rate decoding computes the ratio $\sum_{t=1}^T h_{r, i, t}/T$ of the number of output spikes at any read-out neuron $i$ to example length $T$. Alternatively, time surface decoding  \cite{stewart2021gesture} convolves an exponentially decaying kernel over the time dimension of the read-out spike sequence, and outputs the last sample of the convolution. For the second data set, the real data is rate-encoded as a $T$-length time sequence by sampling from a Bernoulli process with probability $p$ corresponding to the pixel value \cite{gilson2011stdp}. For the third data set, there are exogenous inputs to the generator to encourage diverse samples from the distribution, as we will discuss. 

%These are of the form of a step function $U(\mathrm{t})$ of a given length such that inputs are only spikes where $U(\mathrm{t}) = 1$, with the latency $\mathrm{t}$ sampled at random .  

For the first data set, the discriminator is a $74\times128 \times 1$ feedforward ANN with ReLU activation functions. For the second and third data sets, the discriminator includes several one-dimensional convolutional layers that filter over the time dimension of the data to extract temporal features and learn a natural embedding. The number of layers, as well as the attributes of each layer (number of channels, kernel width, and stride length) are chosen to best match each data set and are detailed in Sec. \ref{sec:resdisc}. The output of the temporal filter is flattened and processed through a linear layer. The approach is adapted from \cite{ko2018deep}. 

\begin{figure}[t!]
\begin{subfigure}{\linewidth}
\begin{minipage}{\linewidth}
    \centering
    \includegraphics[width=\linewidth, trim={0.4in, 0.0in, 0.4in, 0.1in}, clip]{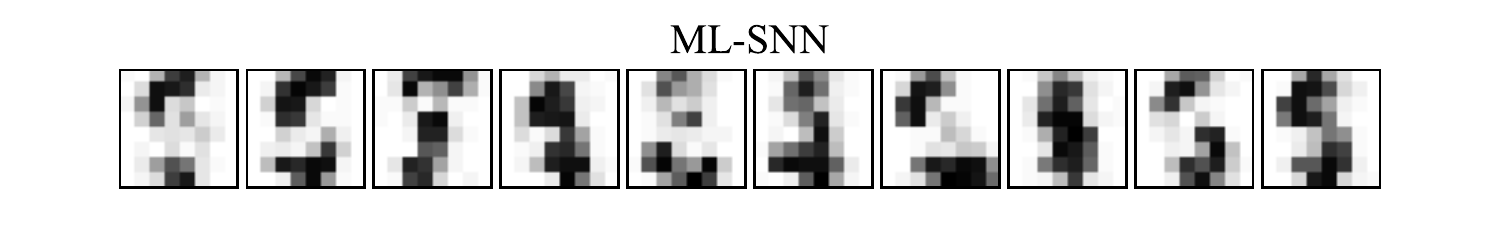}
\end{minipage}
\vfill
\vspace{-0.1in}
\begin{minipage}{\linewidth}
    \centering
    \includegraphics[width=\linewidth, trim={0.4in, 0.1in, 0.4in, 0.1in}, clip]{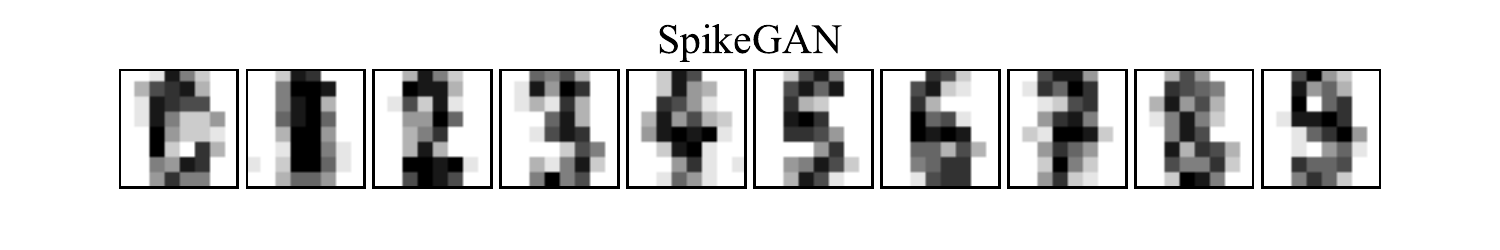}
\end{minipage}
    \caption{}
\end{subfigure}
\vfill
\begin{minipage}{\linewidth}
    \begin{subfigure}{0.45\linewidth}
        \centering
         \includegraphics[width=\linewidth, trim={0.3in, 0.2in, 0.75in, 0.75in}, clip]{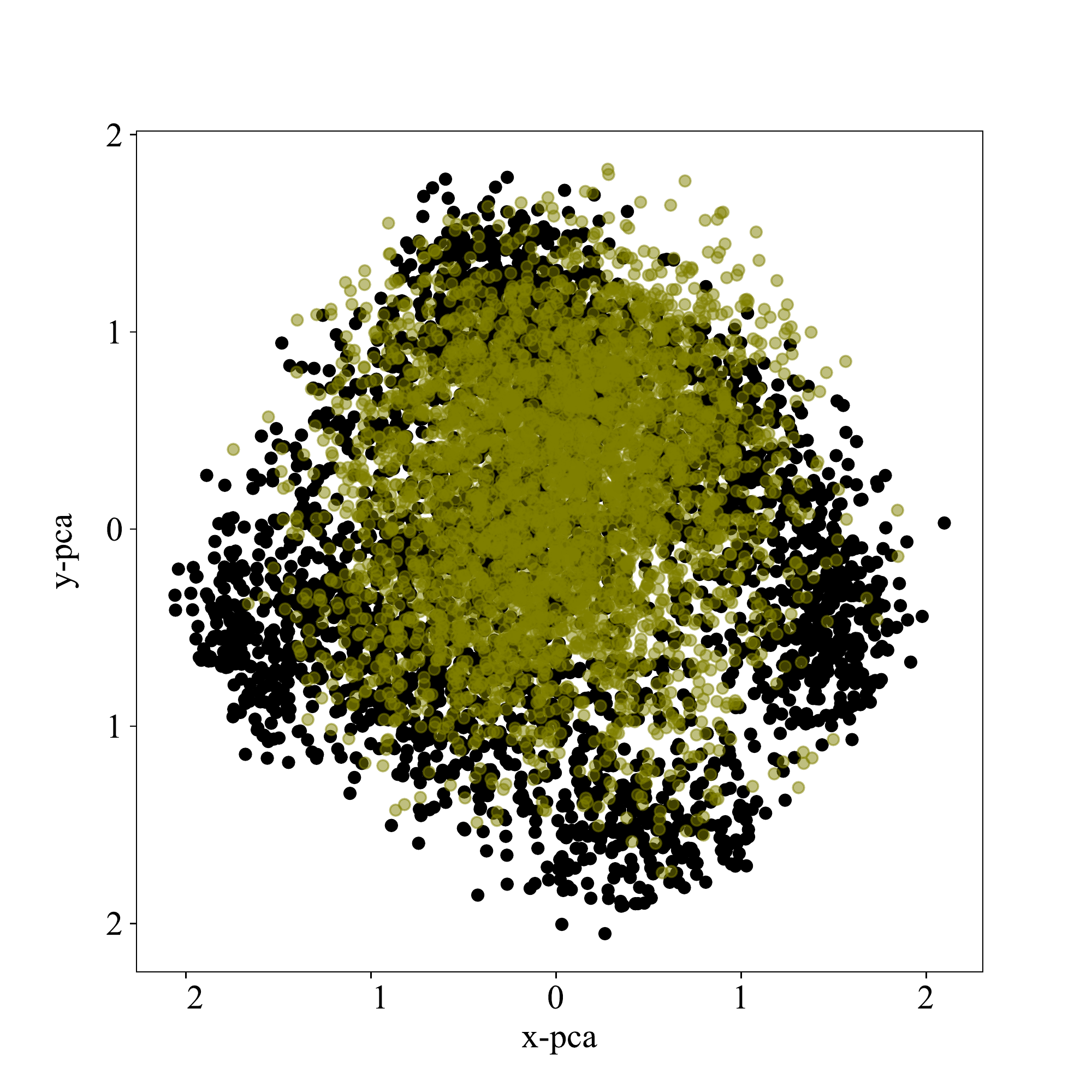}
         \caption{}
    \end{subfigure}
    \hfill
    \begin{subfigure}{0.45\linewidth}
        \centering
         \includegraphics[width=\linewidth, trim={0.3in, 0.2in, 0.75in, 0.75in}, clip]{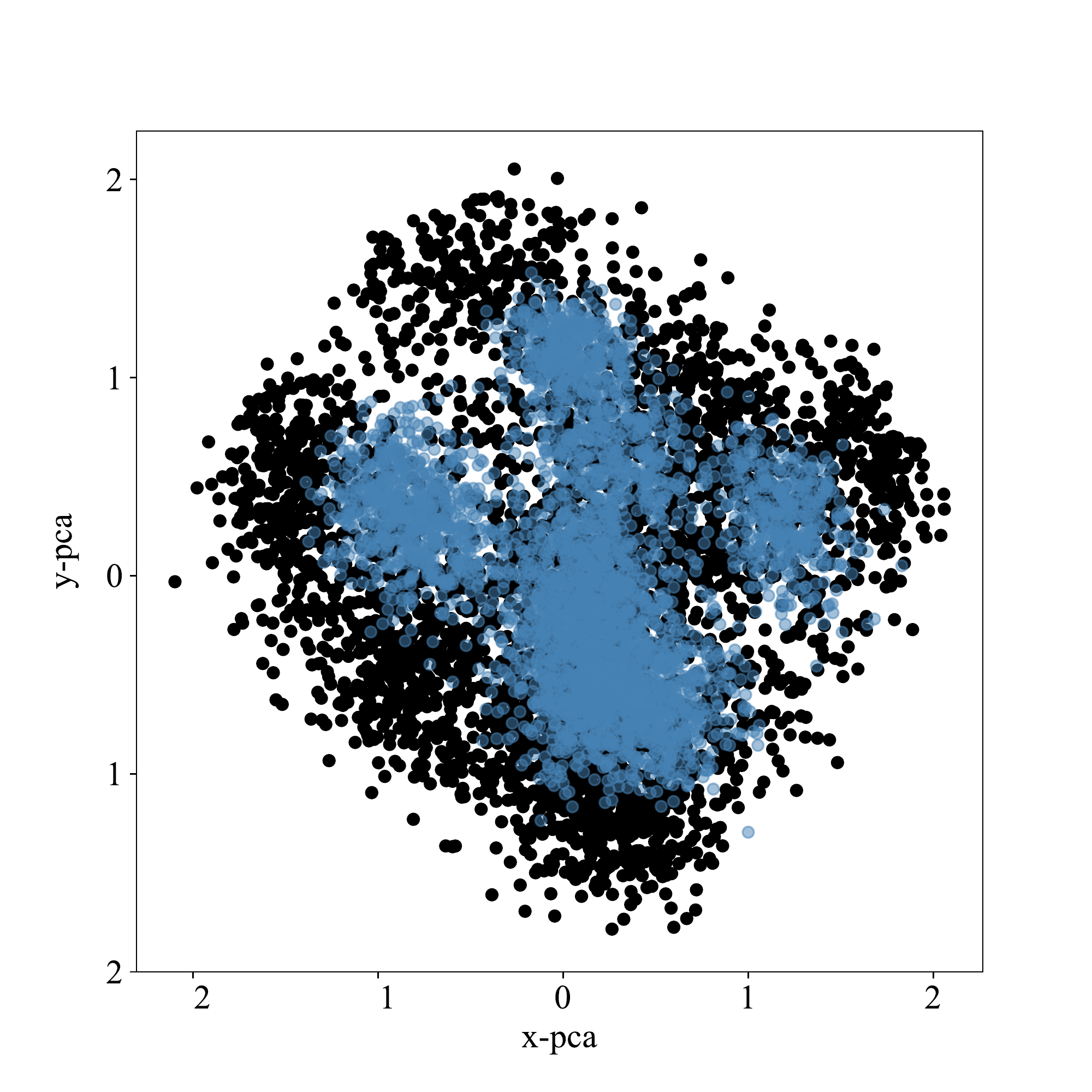}
         \caption{}
    \end{subfigure}
\end{minipage}
\caption{(a) Rate-decoded outputs sampled from an ML-trained SNN \cite{jang2019introduction} (top) and SpikeGAN with a CNN discriminator (bottom). (b, c) PCA projections of a large set of samples drawn from the SpikeGAN and from an ML-SNN respectively (black - real data)}
\label{fig:condconv}
\end{figure}
\subsection{Benchmarks and Performance Metrics}

In order to evaluate how well the output of the SNN generator approximates the underlying population distribution for the data, the following measures are used. 
\begin{enumerate}
\item \emph{Train on synthetic -- Test on real (TSTR) \cite{esteban2017real}}: A classifier is trained over data sampled from the conditional SNN generator for all classes, and test accuracy is evaluated on data sampled from a held-out test set of the real data set. The resulting TSTR error measure provides insight into how well the attributes of the data that are important to distinguish the different data classes have been modeled by the  distribution of the SNN outputs. It specifically captures how fully the SNN samples represent the sample space of the true distribution: If there are outlying portions of the sample space that are not well covered by the sample distribution, the corresponding real data samples may be missclassified, yielding a large TSTR error.

\item \emph{Train on real -- Test on synthetic (TRTS) \cite{esteban2017real}}: A classifier is trained over data from the real data set, and test accuracy is evaluated on synthetic data sampled from the conditional SNN generator for all classes. This measure highlights how well the samples of the synthetic data distribution stay within the bounds of the real data distribution -- or how realistic the samples are. 

\item \emph{Principal component analysis (PCA)}: Extract the principal components of the real data set and compare the projection of a synthetic data set sampled from the SNN generator into that space to the projection of the real data. Plotting the principal component projections gives a visual representation of how well the sample space of the synthetic distribution matches that of the true distribution \cite{esteban2017real}. 
\end{enumerate}

As a benchmark, we consider deep adversarial belief networks (DBNs) \cite{huang2019deep}. DBNs output a single binary sample and hence they can be used only when the data is a vector as for the first data set. They serve as a useful  baseline comparison to the proposed SpikeGAN for the problem of generating real valued handwritten digit images (first data set) in that, like the proposed SNN model, they also implement probabilistic neurons with binary processing capabilities. However, importantly, they lack the capacity to process information encoded over time. 

For temporal real data, i.e., for the second and third data sets, we consider maximum likelihood (ML) learning for a spiking variational auto encoder as detailed in \cite{jang2019introduction} as a benchmark. 

\section{Results and Discussion}\label{sec:resdisc}
In this section, we present our main results by discussing separately the three data sets mentioned in the previous section. We first evaluate single-task performance, and then provide examples also for the continual meta-learning setting. We will mostly focus on the frequentist SpikeGAN approach detailed in Sec. \ref{sec:spikinggan}, but we will also elaborate on the potential advantages of Bayes-SpikeGAN in the context of the third data set.

\begin{figure*}[t!]
\begin{subfigure}{0.5\linewidth}
\begin{minipage}{\linewidth}
    \centering
    \includegraphics[width=\linewidth, trim={0.4in, 0.in, 0.3in, 0.in}, clip]{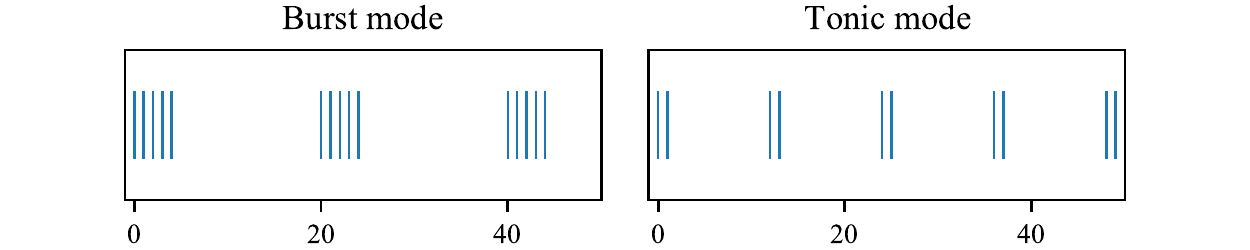}
    \caption{}
    \label{fig:realbursttonic}
\end{minipage}
\vfill
\begin{minipage}{\linewidth}
    \centering
    \includegraphics[width=\linewidth, trim={0.4in, 0.in, 0.3in, 0.in}, clip]{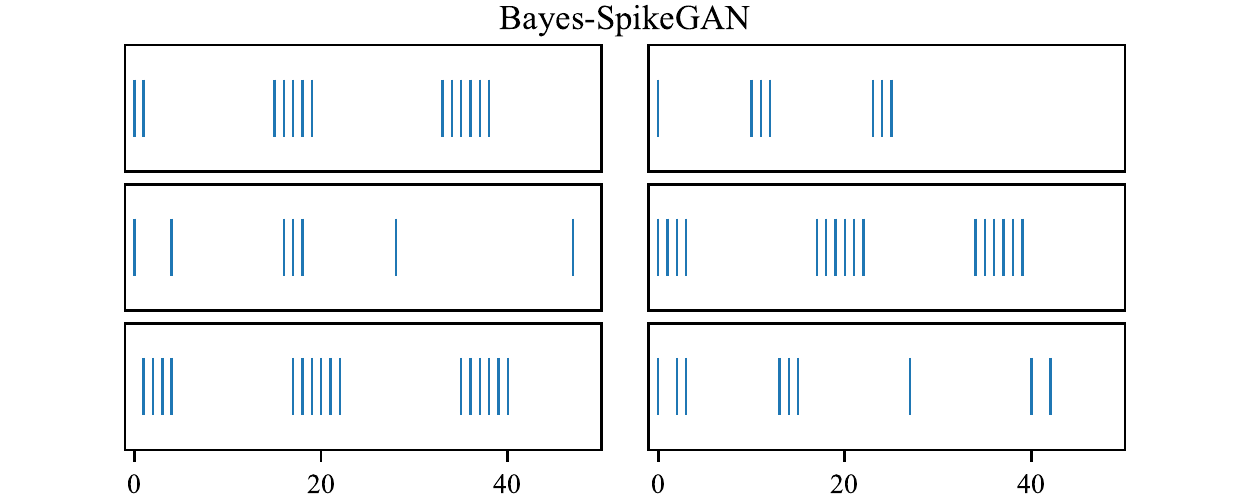}
    \caption{}
    \label{fig:BayesianSpikeGAN}
\end{minipage}
\end{subfigure}
\hfill
\begin{subfigure}{0.5\linewidth}
    \centering
    \includegraphics[width=\linewidth, trim={0.3in, 0.1in, 0.4in, 0.05in}, clip]{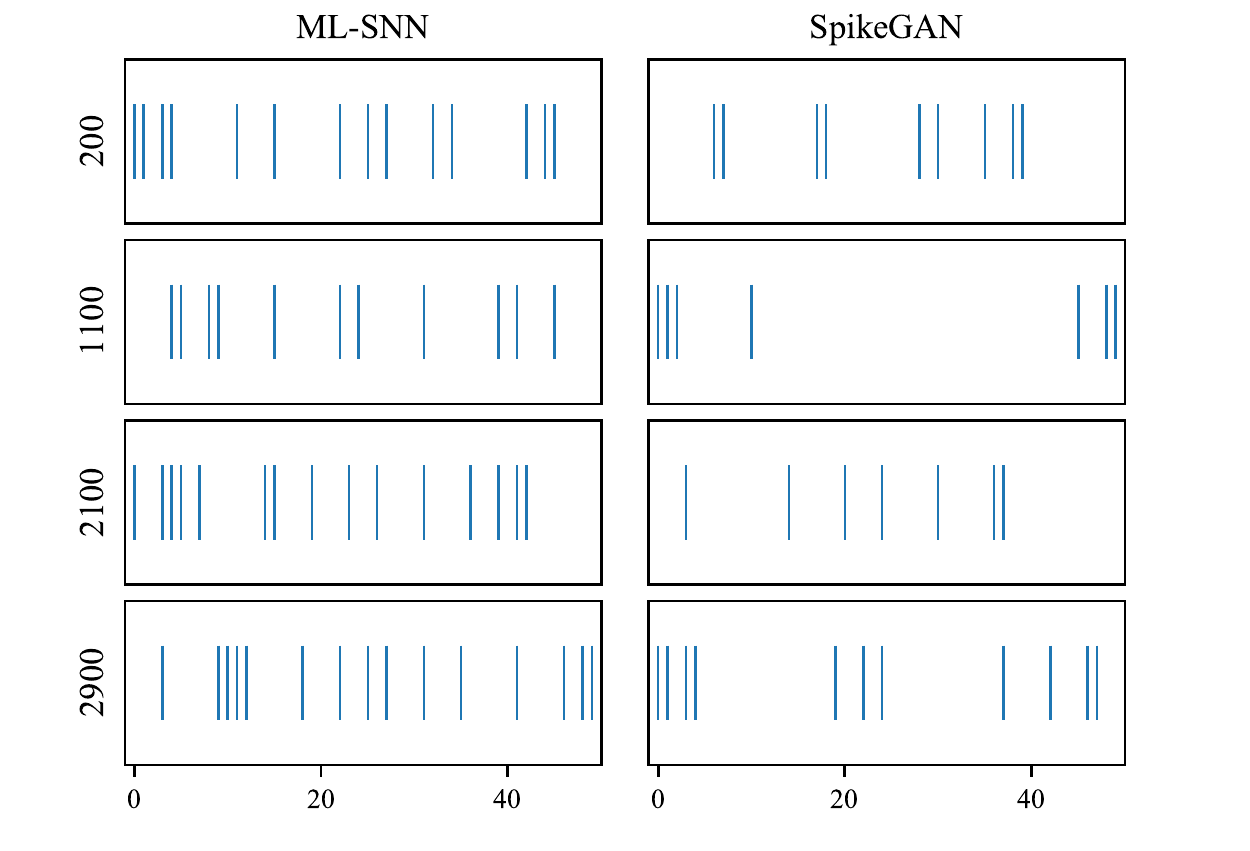}
    \caption{}
    \label{fig:VAE_SpikeGAN_bursttonic}
\end{subfigure}
\caption{(a) `Real' examples of the 2 modes found in the synthetic temporal data set (b) Time sequence samples drawn from the Bayes-SpikeGAN (right) after training is completed. (c) Time sequence samples drawn from the Online VAE (left) and SpikeGAN (right). Samples are taken after the model has been trained for the number of iterations indicated by the label on the left. }
\end{figure*}
\subsection{Handwritten Digits}\label{ssec:handwrittendig}
For this first experiment, the UCI handwritten digits data set \cite{Dua:2019} is considered as defining the real data distribution $p(x|y)$ conditioned on the class label $y$ for $y\in\{0,1,...,9\}$ as a one hot vector. We encode label $y$ as a time sequence input $y_\leq T$ for the SNN using rate encoding. The SNN's task is to learn a temporal distribution $p_{\phi}(x_{\leq T}|y_{\leq T})$ such that the distribution of the synthetic samples $x_{\leq T}$, processed via a fixed decoding scheme,  approximate the samples drawn from the real distribution $p(x|y)$.  As explained in the previous section, both standard rate decoding and time surface decoding are considered for the SNN output sequences $x_{\leq T}$. As an application of the approach, after training, the synthetic data may be considered a temporal representation of the true data and can be used as a neuromorphic data set. 

The real data samples $x$ are $8\times8$ grayscale images with values in the range $[0, 1]$. The SNN generator includes 10 exogenous inputs $y_{\leq T}$, $H_s=128$ supplementary neurons, and $H_r=64$  read-out neurons producing output $x_{\leq T}$; and the SNN has a fully connected topology. An exponential decay basis function $\exp(-\tau/\tau_f)$, with $\tau = 0, 1, ..., \tau_w$, filter length $\tau_w = 5$, and decay rate parameter $\tau_f = 2$, is adopted for both the pre- and post-synaptic filters $\alpha$ and $\beta$ under rate decoding; while a set of two raised cosine basis functions \cite{pillow2008spatio} is used under time surface decoding. 

The TSTR classification accuracy metric is first evaluated by using a $64\times100\times100\times10$ non-spiking ANN classifier with ReLU activation functions that achieves a baseline of 96\% test accuracy when trained on real data, and the results are shown in Fig. \ref{fig:batch100varydecodetrainfake} as a function of the training iterations for the generator. SpikeGAN approaches this ideal accuracy level, while far outperforming the DBN. In this regard, it is noted that, while SpikeGAN can generate grayscale data when paired with a decoder, the DBN can only generate binary data.

% It is also noteworthy that the spiking generator trained using time surface decoding to compress the output sequence performs better than that trained using rate decoding, underscoring the importance of spike timing in training SNNs (Fig. \ref{fig:batch100varydecodetrainfake}). More so, time surface decoding facilitates a solution that requires on average $8.5\%$ fewer read out spikes per image than rate decoding (Fig. \ref{fig:spikecount}) which contributes to a more power efficient model for generating synthetic data online.

 Next we compare the SpikeGAN and DBN GAN in terms of robustness to noise. The noisy data set is constructed by adding uniform noise to a fraction of the pixels in each image selected at random. For the DBN, the images are first binarized as in \cite{huang2019deep} in order to improve the performance, which was otherwise found to be too low in this experiment. The extent to which the digits can be distinguished from the noise in the resulting noisy synthetic images is evaluated using the TRTS accuracy measure. A classifier with the same architecture as in the previous experiment is trained on the uncorrupted real handwritten digit data set and tested on the noisy synthetic data with a baseline comparison to the classifier tested on noisy real data. 
 
 As reported in Fig. \ref{fig:noisedbnvsnn}, the SpikeGAN noisy synthetic data, using time surface decoding, is classified more accurately than the DBN generated noisy synthetic data and maintains an accuracy close to the baseline obtained by testing as the fraction of noisy pixels is increased. This suggests that the capacity of the SNN to generate grayscale images is instrumental in enabling the classifier to distinguish the digits from the noise. This interpretation is corroborated by the samples of synthetic images from SNNs and DBNs shown in Fig. (\ref{fig:noisydata_dbnsnn}). Even for the case of zero pixels with added noise, the SpikeGAN synthetic data is seen to be more realistic than the binary DBN synthetic data. A more quantitative support to this observation is supported by the PCA projections in  Fig. \ref{fig:pca_dbn}, \ref{fig:pca_snntimedecode}. The SpikeGAN projection follows the shape of the data projection, while the DBN projection has many points outside of it.

\subsection{Simulated Neuromorphic Handwritten Digits}
In this second experiment, we move beyond the problem of generating time domain embeddings of real valued data sets by considering the problem of generating synthetic data that matches a spatio-temporal distribution. To simulate a spike domain data set, the UCI handwritten digits data set is encoded via rate encoding to produce the inputs $x_{\leq T}$. The label for each example $y$ that is used as the conditional input, is also encoded using rate encoding as $y_{\leq T}$ before being processed by the discriminator. The discriminator is defined as c128k4s2xc1k4s1x1 (c(number of channels)k(kernel width)s(stride)) with leaky ReLU activation functions, while the SNN generator architecture is the same as in the previous experiment. The key difference between this experiment and the previous is that her the output of the SNN generator is not converted into a real vector before being fed to the discriminator, since the goal is to reproduce the spatio-temporal distribution of the input spiking data set. 

As a benchmark, DBN is not relevant since it cannot generate temporal data, and an SNN trained vial ML as in \cite{jang2019introduction} is used as a reference. Synthetic images are shown by decoding the spiking generator output using the reverse of the encoding scheme applied to the real data, here rate decoding, in Fig. \ref{fig:condconv}, in order to provide a qualitative idea of how well the spatio-temporal distribution has been matched for SpikeGAN and ML training. The PCA projections show that SpikeGAN can represent the mulit-modal structure of the true data distribution more accurately that ML, which is know to be support covering and inclusive \cite{minka2005divergence, li2016r, simeone2017brief}. 

To evaluate the quality of the synthetic data as a neuromorphic data set we now train an SNN classifier using the ML approach in \cite{jang2019introduction} based on the synthetic data, and report the TSTR accuracy metric in \ref{tab:snntrain}. The SNN classifier processes 64 exogenous inputs which are the flattened input image, and includes 256 hidden neurons and 10 visible neurons in a fully connected topology. The visible neurons are clamped to the class labels $y_{\leq T}$ that the synthetic data was conditioned on. The table shows that the SpikeGAN that is trained with a CNN discriminator so that the output directly reproduces a spiking data set enables a better classifier than the SpikeGAN that is trained using a fixed decoder (whether a time surface decoder or rate decoder) to match a real dataset. The VAE does not generate images that match the class label $y_{\leq T}$ that the sample is conditioned on (see Fig. \ref{fig:condconv}) which leads to a poor classifier. The CNN discriminator SpikeGAN approaches the baseline performance reported for the SNN classifier trained over rate encoded real data. \begin{table}[t!]
    \centering
    \caption{TSTR accuracy of SNN classifier trained via ML for simulated neuromorphic handwritten digits}
    \begin{tabular}{l|c}
        Training Data & Real Data Test Accuracy \\
        \hline
         Rate encoded real data & 0.85 \\
         SpikeGAN (CNN discriminator) & 0.82\\
         SpikeGAN time surface decode & 0.8\\
         SpikeGAN rate decode & 0.8\\
         VAE & 0.12
    \end{tabular}
    
    \label{tab:snntrain}
\end{table}
\begin{figure}[t!]
    \centering
    \includegraphics[width=\linewidth, trim={0.2in, 0.1in, 0.5in, 0.4in}, clip]{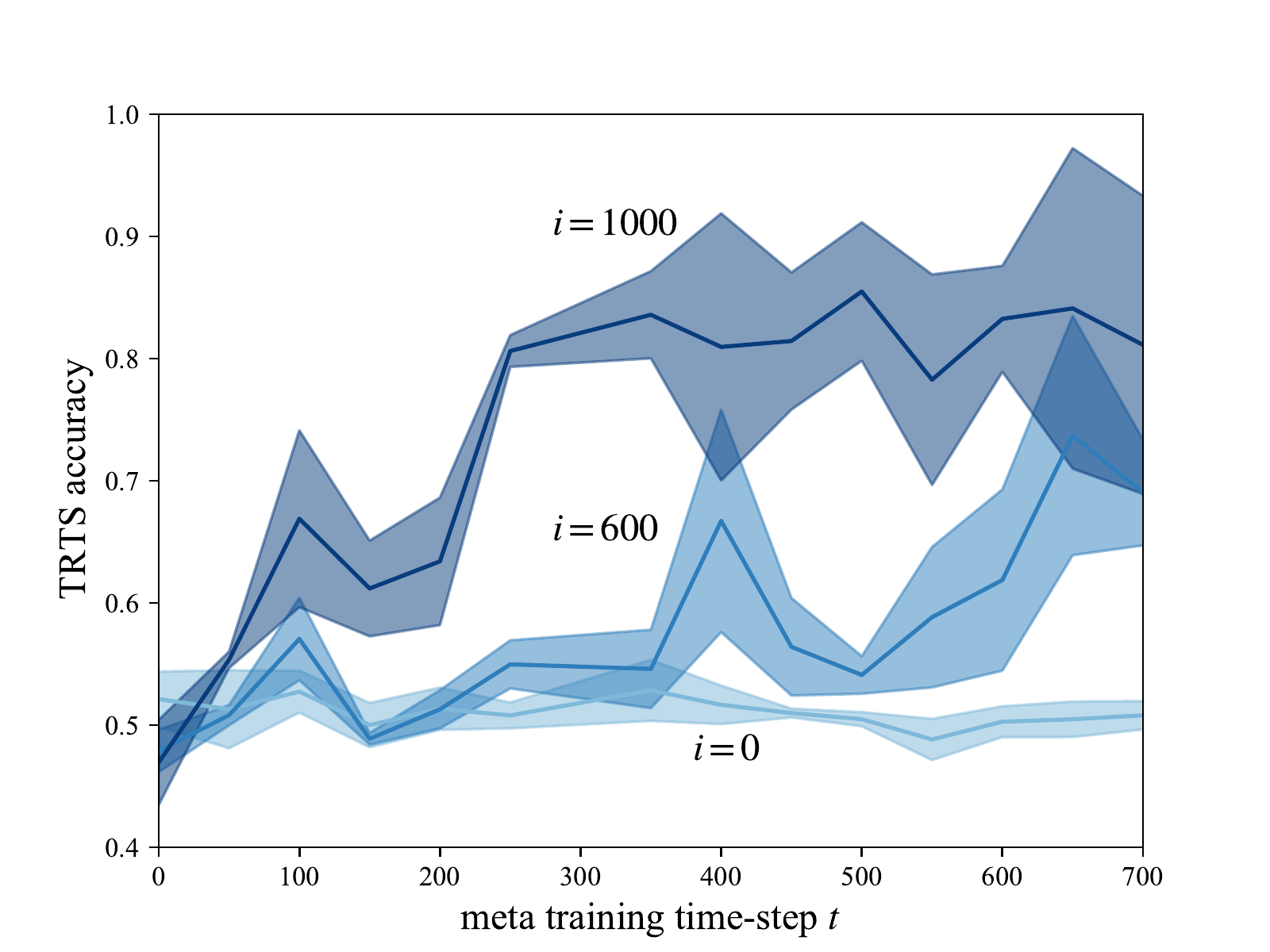}
    \caption{Within task TRTS classification accuracy for hybrid adversarial network pair using the hyperparameter initialization learned over epochs of continual meta training ($t$). Lines are labeled with the number of within-task adversarial training iterations ($i$), and show the average over three tasks drawn from a held out set of digits (shading shows half standard deviation spread).}
    \label{fig:metaadversarial}
\end{figure}

\subsection{Synthetic Temporal Data}
For the last SpikeGAN experiment, a synthetic temporal data set is constructed by taking inspiration from biological neuronal behavior \cite{weber2017capturing}. The goal is to assess whether adversarial unsupervised training can reproduce some the diversity shown by neuronal activity in the brain. We specifically consider two biologically inspired neural spike modes, namely tonic spiking and burst spiking. In a manner similar to \cite{weber2017capturing}, we define burst spiking as periods of 5 consecutive spikes followed by a non-spiking period of 15 time steps while tonic spiking includes 2 consecutive spikes with a 10 time step non-spiking period as displayed in Fig. \ref{fig:realbursttonic}. 

As shown in \cite{weber2017capturing}, a single neuron with tailored synaptic filters is sufficient to approximate either one of these modes well. Both the synaptic filter and spiking threshold (bias) of the neuron need to be carefully optimized to maintain this behavior. We generate a data set of 10,000 burst and tonic spiking sequences of length $T=50$. We compare the performance of the SpikeGAN, and Bayes-SpikeGAN with an SNN trained via ML as the baseline. Specifically, we train the ML-based SNN and the SpikeGAN each with a single neuron with $T=50$ and synaptic filter memory $\tau_w=30$ that is stimulated by an exogenous step function input, as well as Bayes-SpikeGAN with $J = 5$ of the single neuron SNN generators to approximate the posterior distribution over $\phi$. For the Bayes-SpikeGAN we make the choice an improper constant prior for $p(\phi)$ in the SVGD update rule (\ref{eq:svgd_update}).  

As seen in Fig. \ref{fig:VAE_SpikeGAN_bursttonic}, the ML-trained SNN generates outputs that are a blend of the two modes while the SpikeGAN oscillates between them as training proceeds. In contrast, as seen in Fig. \ref{fig:BayesianSpikeGAN}, Bayes-SpikeGAN is able to learn a set of generators whose combined outputs cover both modes simultaneously. 

% The train on real, test on synthetic accuracy is evaluated on a convolutional classifier network with the same architecture as the discriminator (changing only the number of read-out neurons). The distribution of class labels attributed to the synthetic data by the classifier is an indication of whether the synthetic data covers all modes of the data. 

\subsection{Continual Meta-Learning}
We now evaluate the ability of meta-SpikeGAN to continually improve its efficiency in generating useful time domain embeddings by focusing on the real-valued handwritten digits data set studied in Sec. \ref{ssec:handwrittendig}. The real data set that defines each task $T^{(t)}$ is chosen as the subset of the UCI handwritten digits data set obtained by selecting the combination of two digits from among digits 0-6, along with a randomly sampled rotation of $90^{\circ}$ applied to each digit. Digits 7-9 are reserved for testing. The class labels $y\in\{0, 1\}$ are applied to the pair of rotated digits in each new task. As in the previous handwritten digits experiment, the SNN generator is conditioned on the class label as a one hot vector encoded as time sequence $y_{\leq T}$ using rate encoding with $T=5$. The output of the SNN generator $x_{\leq T}$ is decoded back to a natural signal using rate decoding before being fed to the discriminator. 

The SNN generator includes two exogenous inputs $y_{\leq T}$, $H_s = 100$ supplementary neurons and $H_r=64$ read-out neurons producing output $x_{\leq T}$ and has a fully connected topology. The same exponential decay synaptic filters are used as described in Sec. \ref{ssec:handwrittendig}. We implement the $\MetaUpdate(\boldsymbol{\theta}^{(t,i)}, \{D^{(n)}\}_{n=1}^N)$ function with $N=10$ within-task data sets of $M=5$ examples each and 10 within-task update steps. 

The performance of SpikeGAN under the meta-SpikeGAN initialization $\boldsymbol{\theta}^{(t, i)} = (\Theta^{(t,i)}, \theta^{(t, i)})$ is evaluated by looking at the TRTS accuracy for synthetic data generated at intervals throughout within-task training. If training efficiency has been improved by the meta-SpikeGAN initialization, the TRTS accuracy will be higher after fewer within-task training updates $i$. The continual improvement of the meta-SpikeGAN hyperparameter initialization is measured by applying the initialization to a SpikeGAN model after every 50 meta-training time-steps $t$ and by evaluating the TRTS accuracy throughout within-task training on a new task. 

We choose a new task as the combination of two digits from the set of held-out digits (digits 7-9) of the UCI handwritten digits data set and train the SpikeGAN over mini-batches of $B=100$ training examples. As shown in Fig. \ref{fig:metaadversarial}, the TRTS accuracy improves significantly as the meta-SpikeGAN hyperparameter initialization is learned, with the accuracy after $i=1000$ within-task updates increasing by 30\% over a randomly initialized SpikeGAN ($t=0$ meta training time-steps) as meta training progresses.

\section{Conclusion}
This paper has introduced adversarial training methods for a novel hybrid SNN-ANN GAN architecture, termed SpikeGAN. The proposed approaches solve the problem of learning how to emulate a spatio-temporal distribution, while allowing for a flexible, distribution-based, definition of the target outputs that fully leverages the temporal encoding nature of spiking signals. Both frequentist and Bayesian formulations of the learning problem were considered, along with a generalization to continual meta-learning. The proposed SpikeGAN approach has been evaluated on a range of spatio-temporal data sets, and has been shown to outperform current baselines (DBN GANs and SNNs based on ML training) in all settings. Bayes-SpikeGAN is proven to be an important extension to the frequentist learning solution in the problem of emulating multi-modal data with large variations in specific temporal patterns, such as for biologically inspired spiking sequences. We leave as future work the problem of designing online learning rules for SpikeGANs that leverage a recurrent network as the discriminator.

% or


\appendices
\section{SNN Maximum Likelihood Optimization}\label{app:maxlikelihood}
The expected log-likelihood of the observed spikes is estimated via Monte Carlo sampling of the hidden spikes $[h_{i, \tau}\sim p_{\phi}(h_{i,\tau}| u_{i, \tau})]_{i\in\mathcal{H}}$ at every discrete time-instant $\tau$ which allows the computation of the new membrane potentials $[u_{i,\tau+1}]_{i\in\mathcal{V},\mathcal{H}}$ according to Eq. (\ref{eq:glm_mempot}). For the visible neurons, the model parameters are iteratively updated according to the local gradient (Eq. \ref{eq:grad_local}). For the hidden neurons, the online gradient is estimated by the REINFORCE gradient \cite{jang2019introduction}
\begin{equation}\label{eq:grad_hidden}
    \nabla_{\theta_{i\in\mathcal{H}}} L_{\upsilon_{\leq \mathcal{T}}}(\theta) \simeq \sum_{\tau=1}^{\mathcal{T}} \ell_{\tau} \nabla_{\theta} \log p(h_{\tau}| u_{\leq\tau-1})
\end{equation} where $\nabla_{\theta} \log p(h_{\tau}| u_{\leq\tau})$ is evaluated as in (Eq. \ref{eq:grad_local}), substituting $h_{i,\tau}$ for $\upsilon_{i,\tau}$, and we have the global error, or learning signal
\begin{equation}\label{eq:learning_sig}
    \ell_{\tau} = \sum_{i\in\mathcal{V}}\log\left(\overline{\upsilon}_{i,\tau}\overline{\sigma}(u_{i,\tau})+ \upsilon_{i,\tau}\sigma(u_{i,\tau}) \right).
\end{equation}
This maximum likelihood training update may also include a sparsity-inducing regularizer as described in \cite{jang2019introduction}.

%\section*{Acknowledgment}


\ifCLASSOPTIONcaptionsoff
  \newpage
\fi



\bibliographystyle{IEEEtran}
\bibliography{IEEEabrv, references}

\begin{thebibliography}{10}
\providecommand{\url}[1]{#1}
\csname url@samestyle\endcsname
\providecommand{\newblock}{\relax}
\providecommand{\bibinfo}[2]{#2}
\providecommand{\BIBentrySTDinterwordspacing}{\spaceskip=0pt\relax}
\providecommand{\BIBentryALTinterwordstretchfactor}{4}
\providecommand{\BIBentryALTinterwordspacing}{\spaceskip=\fontdimen2\font plus
\BIBentryALTinterwordstretchfactor\fontdimen3\font minus
  \fontdimen4\font\relax}
\providecommand{\BIBforeignlanguage}[2]{{%
\expandafter\ifx\csname l@#1\endcsname\relax
\typeout{** WARNING: IEEEtran.bst: No hyphenation pattern has been}%
\typeout{** loaded for the language `#1'. Using the pattern for}%
\typeout{** the default language instead.}%
\else
\language=\csname l@#1\endcsname
\fi
#2}}
\providecommand{\BIBdecl}{\relax}
\BIBdecl

\bibitem{stewart2020chip}
K.~Stewart, G.~Orchard, S.~B. Shrestha, and E.~Neftci, ``On-chip few-shot
  learning with surrogate gradient descent on a neuromorphic processor,'' in
  \emph{Proc. International Conference on Artificial Intelligence Circuits and
  Systems (AICAS)}.\hskip 1em plus 0.5em minus 0.4em\relax IEEE, 2020, pp.
  223--227.

\bibitem{huh2017gradient}
D.~Huh and T.~J. Sejnowski, ``Gradient descent for spiking neural networks,''
  \emph{arXiv preprint arXiv:1706.04698}, 2017.

\bibitem{jimenez2014stochastic}
D.~Jimenez~Rezende and W.~Gerstner, ``Stochastic variational learning in
  recurrent spiking networks,'' \emph{Frontiers in computational neuroscience},
  vol.~8, p.~38, 2014.

\bibitem{jang2019introduction}
H.~Jang, O.~Simeone, B.~Gardner, and A.~Gruning, ``An introduction to
  probabilistic spiking neural networks: Probabilistic models, learning rules,
  and applications,'' \emph{IEEE Signal Processing Magazine}, vol.~36, no.~6,
  pp. 64--77, 2019.

\bibitem{serrano2013128}
T.~Serrano-Gotarredona and B.~Linares-Barranco, ``A 128$\times$128 1.5\%
  contrast sensitivity 0.9\% fpn 3 $\mu$s latency 4 mw asynchronous frame-free
  dynamic vision sensor using transimpedance preamplifiers,'' \emph{IEEE
  Journal of Solid-State Circuits}, vol.~48, no.~3, pp. 827--838, 2013.

\bibitem{liu2013asynchronous}
S.-C. Liu, A.~van Schaik, B.~A. Minch, and T.~Delbruck, ``Asynchronous binaural
  spatial audition sensor with 2$\times$64$\times$4 channel output,''
  \emph{IEEE transactions on biomedical circuits and systems}, vol.~8, no.~4,
  pp. 453--464, 2013.

\bibitem{pan2019neural}
Z.~Pan, J.~Wu, M.~Zhang, H.~Li, and Y.~Chua, ``Neural population coding for
  effective temporal classification,'' in \emph{2019 International Joint
  Conference on Neural Networks (IJCNN)}.\hskip 1em plus 0.5em minus
  0.4em\relax IEEE, 2019, pp. 1--8.

\bibitem{gui2020review}
J.~Gui, Z.~Sun, Y.~Wen, D.~Tao, and J.~Ye, ``A review on generative adversarial
  networks: Algorithms, theory, and applications,'' \emph{arXiv preprint
  arXiv:2001.06937}, 2020.

\bibitem{saxena2021generative}
D.~Saxena and J.~Cao, ``Generative adversarial networks (gans) challenges,
  solutions, and future directions,'' \emph{ACM Computing Surveys (CSUR)},
  vol.~54, no.~3, pp. 1--42, 2021.

\bibitem{pillow2008spatio}
J.~W. Pillow, J.~Shlens, L.~Paninski, A.~Sher, A.~M. Litke, E.~Chichilnisky,
  and E.~P. Simoncelli, ``Spatio-temporal correlations and visual signalling in
  a complete neuronal population,'' \emph{Nature}, vol. 454, no. 7207, pp.
  995--999, 2008.

\bibitem{saatci2017bayesian}
Y.~Saatci and A.~Wilson, ``Bayesian gans,'' in \emph{Advances in neural
  information processing systems}, 2017, pp. 3624--3633.

\bibitem{weber2017capturing}
A.~I. Weber and J.~W. Pillow, ``{Capturing the Dynamical Repertoire of Single
  Neurons with Generalized Linear Models},'' \emph{Neural Computation},
  vol.~29, no.~12, pp. 3260--3289, 2017.

\bibitem{wang2020meta}
D.~Wang, B.~Ding, and D.~Feng, ``Meta reinforcement learning with generative
  adversarial reward from expert knowledge,'' in \emph{2020 IEEE 3rd
  International Conference on Information Systems and Computer Aided Education
  (ICISCAE)}.\hskip 1em plus 0.5em minus 0.4em\relax IEEE, 2020, pp. 1--7.

\bibitem{kotariya2021spiking}
V.~Kotariya and U.~Ganguly, ``Spiking-gan: A spiking generative adversarial
  network using time-to-first-spike coding,'' \emph{arXiv preprint
  arXiv:2106.15420}, 2021.

\bibitem{singh2020nebula}
S.~Singh, A.~Sarma, N.~Jao, A.~Pattnaik, S.~Lu, K.~Yang, A.~Sengupta,
  V.~Narayanan, and C.~R. Das, ``Nebula: a neuromorphic spin-based ultra-low
  power architecture for snns and anns,'' in \emph{2020 ACM/IEEE 47th Annual
  International Symposium on Computer Architecture (ISCA)}.\hskip 1em plus
  0.5em minus 0.4em\relax IEEE, 2020, pp. 363--376.

\bibitem{pei2019towards}
J.~Pei, L.~Deng, S.~Song, M.~Zhao, Y.~Zhang, S.~Wu, G.~Wang, Z.~Zou, Z.~Wu,
  W.~He \emph{et~al.}, ``Towards artificial general intelligence with hybrid
  tianjic chip architecture,'' \emph{Nature}, vol. 572, no. 7767, pp. 106--111,
  2019.

\bibitem{yang2019dashnet}
Z.~Yang, Y.~Wu, G.~Wang, Y.~Yang, G.~Li, L.~Deng, J.~Zhu, and L.~Shi,
  ``Dashnet: A hybrid artificial and spiking neural network for high-speed
  object tracking,'' \emph{arXiv preprint arXiv:1909.12942}, 2019.

\bibitem{stewart2021gesture}
K.~Stewart, A.~Danielescu, L.~Supic, T.~Shea, and E.~Neftci, ``Gesture
  similarity analysis on event data using a hybrid guided variational auto
  encoder,'' \emph{arXiv preprint arXiv:2104.00165}, 2021.

\bibitem{skatchkovsky2021learning}
N.~Skatchkovsky, O.~Simeone, and H.~Jang, ``Learning to time-decode in spiking
  neural networks through the information bottleneck,'' \emph{in Proc.
  NeurIPS}, 2021.

\bibitem{rosenfeld2021fast}
B.~Rosenfeld, B.~Rajendran, and O.~Simeone, ``Fast on-device adaptation for
  spiking neural networks via online-within-online meta-learning,'' in
  \emph{Data Science and Learning Workshop}, 2021.

\bibitem{goodfellow2014generative}
I.~Goodfellow, J.~Pouget-Abadie, M.~Mirza, B.~Xu, D.~Warde-Farley, S.~Ozair,
  A.~Courville, and Y.~Bengio, ``Generative adversarial nets,'' \emph{Advances
  in neural information processing systems}, vol.~27, 2014.

\bibitem{fremaux2016neuromodulated}
N.~Fr{\'e}maux and W.~Gerstner, ``Neuromodulated spike-timing-dependent
  plasticity, and theory of three-factor learning rules,'' \emph{Frontiers in
  neural circuits}, vol.~9, p.~85, 2016.

\bibitem{yoon2019time}
J.~Yoon, D.~Jarrett, and M.~Van~der Schaar, ``Time-series generative
  adversarial networks,'' 2019.

\bibitem{liu2016stein}
Q.~Liu and D.~Wang, ``Stein variational gradient descent: A general purpose
  bayesian inference algorithm,'' \emph{arXiv preprint arXiv:1608.04471}, 2016.

\bibitem{finn2019online}
C.~Finn, A.~Rajeswaran, S.~Kakade, and S.~Levine, ``Online meta-learning,''
  \emph{arXiv preprint arXiv:1902.08438}, 2019.

\bibitem{nichol2018first}
A.~Nichol, J.~Achiam, and J.~Schulman, ``On first-order meta-learning
  algorithms,'' \emph{arXiv preprint arXiv:1803.02999}, 2018.

\bibitem{huang2019deep}
Y.~Huang, A.~Panahi, H.~Krim, Y.~Yu, and S.~L. Smith, ``Deep adversarial belief
  networks,'' \emph{arXiv preprint arXiv:1909.06134}, 2019.

\bibitem{Dua:2019}
\BIBentryALTinterwordspacing
D.~Dua and C.~Graff, ``{UCI} machine learning repository,'' 2017. [Online].
  Available: \url{http://archive.ics.uci.edu/ml}
\BIBentrySTDinterwordspacing

\bibitem{gilson2011stdp}
M.~Gilson, T.~Masquelier, and E.~Hugues, ``Stdp allows fast rate-modulated
  coding with poisson-like spike trains,'' \emph{PLoS computational biology},
  vol.~7, no.~10, p. e1002231, 2011.

\bibitem{ko2018deep}
W.~Ko, J.~Yoon, E.~Kang, E.~Jun, J.-S. Choi, and H.-I. Suk, ``Deep recurrent
  spatio-temporal neural network for motor imagery based bci,'' in \emph{2018
  6th International Conference on Brain-Computer Interface (BCI)}.\hskip 1em
  plus 0.5em minus 0.4em\relax IEEE, 2018, pp. 1--3.

\bibitem{esteban2017real}
C.~Esteban, S.~L. Hyland, and G.~R{\"a}tsch, ``Real-valued (medical) time
  series generation with recurrent conditional gans,'' \emph{arXiv preprint
  arXiv:1706.02633}, 2017.

\bibitem{minka2005divergence}
T.~Minka \emph{et~al.}, ``Divergence measures and message passing,'' Citeseer,
  Tech. Rep., 2005.

\bibitem{li2016r}
Y.~Li and R.~E. Turner, ``R\'enyi divergence variational inference,''
  \emph{arXiv preprint arXiv:1602.02311}, 2016.

\bibitem{simeone2017brief}
O.~Simeone, ``A brief introduction to machine learning for engineers,''
  \emph{Foundations and Trends{\textregistered} in Signal Processing}, vol.~12,
  no. 3-4, pp. 200--431, 2018.

\end{thebibliography}
\end{document}